\documentclass[10pt]{article} %
\usepackage[accepted]{tmlr}

\usepackage{amsmath,amsfonts,bm}

\def\eqref#1{equation~\ref{#1}}

\def\1{\bm{1}}

\DeclareMathAlphabet{\mathsfit}{\encodingdefault}{\sfdefault}{m}{sl}
\SetMathAlphabet{\mathsfit}{bold}{\encodingdefault}{\sfdefault}{bx}{n}

\usepackage{hyperref}
\usepackage{url}

\usepackage[dvipsnames]{xcolor}
\usepackage{graphicx}
\usepackage{xspace}
\usepackage{caption}
\usepackage{tikz}
\usepackage{pgfplots}
\usepackage{wrapfig}
\usepackage{enumitem}
\usepackage{subcaption}
\usepackage{colortbl} %
\usepackage{booktabs}
\newcommand{\modelname}{ReVision\@\xspace}
\newcommand{\motionmodelname}{PMP\@\xspace}
\newcommand{\eg}{\textit{e.g.}\@\xspace}
\newcommand{\ie}{\textit{i.e.}\@\xspace}
\newcommand{\etal}{\textit{et al.}\@\xspace}

\title{\modelname: Refining Video Diffusion with Explicit 3D Motion Modeling}

\author{\name Qihao Liu \email qliu45@jhu.edu \\
      \addr Johns Hopkins University
      \AND
      \name Ju He \email jhe47@jhu.edu \\
      \addr Johns Hopkins University
      \AND
      \name Qihang Yu \email yucornetto@gmail.com \\
      \addr Independent Researcher
      \AND
      \name Liang-Chieh Chen\footnotemark[2] \email lcchen@cs.ucla.edu\\
      \addr Independent Researcher
      \AND
      \name Alan Yuille\footnotemark[2] \email ayuille1@jhu.edu\\
      \addr Johns Hopkins University
      }

\def\month{01}  %
\def\year{2026} %
\def\openreview{\url{https://openreview.net/forum?id=mQ5frFQTFV}} %

\begin{document}

\maketitle
\footnotetext[2]{Equal advising.}
\vspace{-1mm}
\begin{abstract}
In recent years, video generation has seen significant advancements.
However, challenges still persist in generating complex motions and interactions. 
To address these challenges, we introduce \modelname, a plug-and-play framework that explicitly integrates parameterized 3D model knowledge into a pretrained conditional video generation model, significantly enhancing its ability to generate high-quality videos with complex motion and interactions.
Specifically, \modelname consists of three stages.
First, a video diffusion model is used to generate a coarse video.
Next, we extract a set of 2D and 3D features from the coarse video to construct a 3D object-centric representation, which is then refined by our proposed parameterized motion prior model to produce an accurate 3D motion sequence.
Finally, this refined motion sequence is fed back into the same video diffusion model as additional conditioning, enabling the generation of motion-consistent videos, even in scenarios involving complex actions and interactions.
We validate the effectiveness of our approach on Stable Video Diffusion, where \modelname significantly improves motion fidelity and coherence.
Remarkably, with only 1.5B parameters, it even outperforms a state-of-the-art video generation model with over 13B parameters on complex video generation by a substantial margin.
Our results suggest that, by incorporating 3D motion knowledge, even a relatively small video diffusion model can generate complex motions and interactions with greater realism and controllability, offering a promising solution for physically plausible video generation. 
Project page:~\url{https://revision-video.github.io/}
\end{abstract}
\vspace{-1mm}

\begin{figure}[htbp]
    \centering
    \captionsetup{type=figure}
    \vspace{-3mm}
    \includegraphics[width=\linewidth]{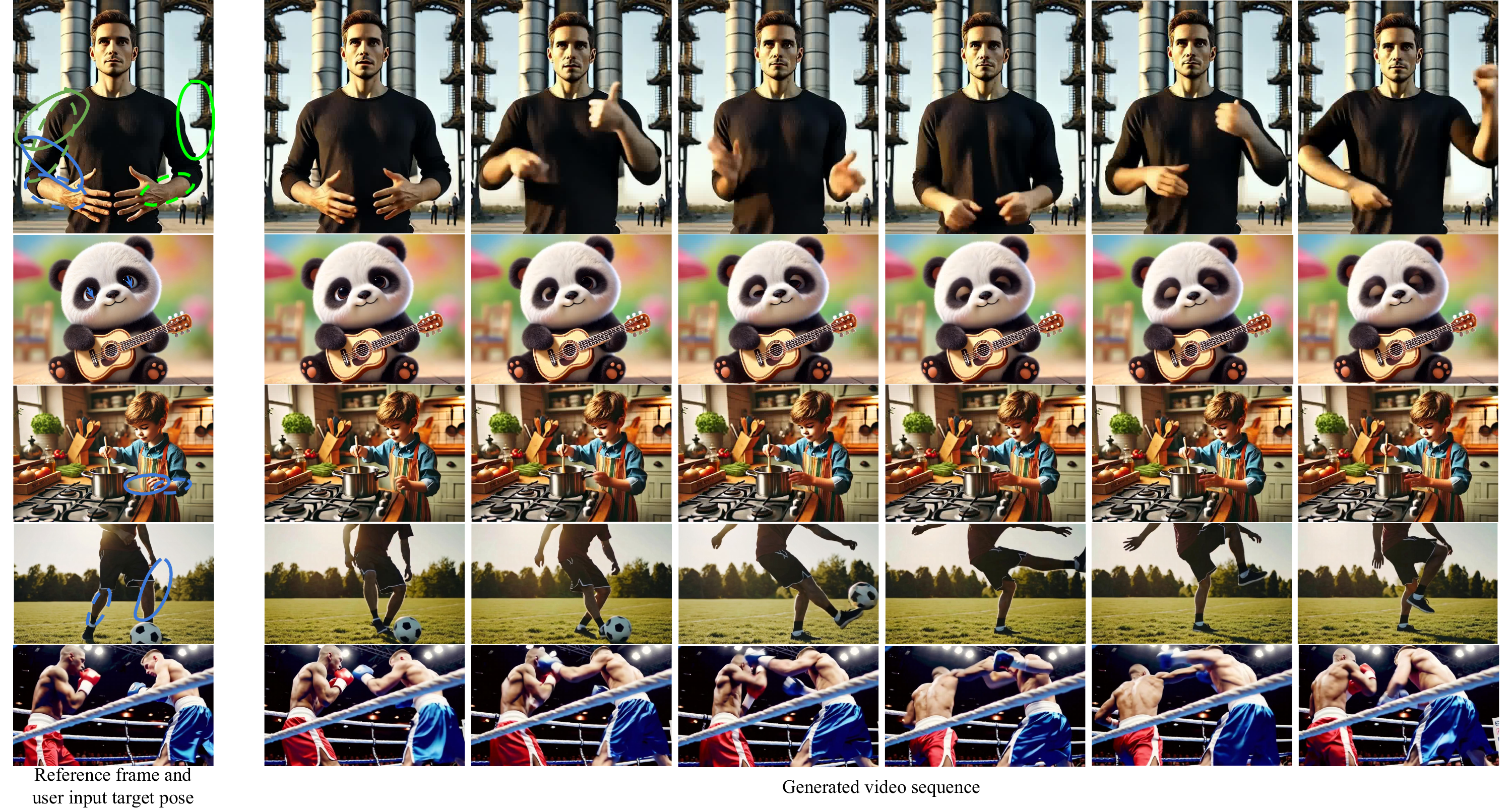}
    \vspace{-3mm}
    \captionof{figure}{
    By explicitly leveraging a parameterized 3D motion model, \modelname enhances pre-trained video generation models (\eg, Stable Video Diffusion) to produce high-quality videos with complex motion (row 1), enabling precise motion control (rows 2, 3) and accurate interactions (rows 4, 5).
    During inference, an \textit{optional} target pose can be specified via a rough sketch (rows 1, 3, 4, colored circles for different parts, dashed lines for the original pose, solid lines for the target pose) or a simple drag operation (\textcolor{blue}{blue} arrows in row 2) indicating the final position.
    }
    \vspace{-2mm}
    \label{fig:teaser}
\end{figure}
    
\section{Introduction}
\label{sec:intro}

Video diffusion models have achieved remarkable success in producing high-quality, temporally coherent videos~\citep{blattmann2023stable, brooks2024video, polyak2024moviegencastmedia, kong2024hunyuanvideo}. 
It has been driven by advances in model architectures~\citep{peebles2023scalable}, increases in model complexity, reaching tens of billion parameters~\citep{polyak2024moviegencastmedia}, and the availability of large-scale high-quality datasets~\citep{chen2024panda}.
However, current models still struggle to generate videos that adhere to realistic physical principles, making it difficult to consistently achieve fine-grained motion control, complex movements, and coherent object interactions.
Despite extensive efforts to improve performance through larger models and higher-quality datasets, a recent study~\citep{kang2024far} indicates that scaling model size and data alone is insufficient to fully capture the complexities of the real world.

On the other hand, human image animation models~\citep{hu2024animate, tan2024animate, xu2024magicanimate} offer valuable insights for addressing persistent challenges in video generation.
Despite using smaller models and less data, these methods achieve consistent and precise video outputs with complex motions by following predefined 2D keypoint trajectories.
This success suggests that incorporating a well-defined motion prior can substantially reduce the learning complexity of video generative models, enabling them to generate coherent and lifelike motion.
However, in general video generation tasks, such strong guiding signals are typically unavailable, limiting the direct applicability of these animation techniques to broader video generation scenarios.
This raises a critical question: \textit{Can we develop a video generation framework that leverages the implicit motion information embedded in the generated videos as guidance to further enhance video quality?}

In this paper, we propose a simple, general, and plug-and-play video generation framework that incorporates motion knowledge into a conditional video generation model via a parameterized 3D representation, allowing the generation of videos with complex motions and interactions involving \textit{humans}, \textit{animals}, and \textit{general objects}.
The core of \textbf{\modelname} is to \textbf{Re}generate \textbf{Vi}deos with explicit 3D motion representat\textbf{ion}s, following an \textit{Extract–Optimize–Reinforce} pipeline.
Specifically, to effectively leverage 3D knowledge without heavy retraining of the diffusion model while preserving its original ability to generate high-quality visual appearance, we design the pipeline in three stages.

In the first stage, we employ a video diffusion model, \eg, SVD~\citep{blattmann2023stable}, to generate a coarse video conditioned on the given input.
In the second stage, we utilize parametric 3D models (\ie, SMPL-X~\citep{pavlakos2019expressive} for humans, SMAL~\citep{rueegg2023bite, Zuffi:CVPR:2024} for animals, and 2D binary mask~\citep{yu2023convolutions} with estimated depth~\citep{depth_anything_v2} for general objects) to \textit{extract} 3D shape and motion features from the coarse video.
These 3D object-centric representations are subsequently \textit{optimized} by the proposed \textbf{P}arameterized \textbf{M}otion \textbf{P}rior model (\textbf{\motionmodelname}), producing a more accurate and natural 3D motion sequence.
In the third stage, the refined 3D motion sequences are incorporated as additional conditioning inputs to \textit{reinforce} the diffusion model, enabling it to regenerate the video with improved coherence and realism.

Extensive qualitative results and human preference studies confirm that our model excels at generating complex motions and interactions.
We first apply \modelname on Stable Video Diffusion (SVD)~\citep{blattmann2023stable}, substantially improving its ability to generate realistic and intricate motions.
We further compare \modelname-SVD with HunyuanVideo~\citep{kong2024hunyuanvideo}, a state-of-the-art open-source video generation model with 13B parameters, and demonstrate superior motion quality.
Finally, on the particularly challenging dance generation task, our model outperforms state-of-the-art human image animation methods that rely on ground-truth pose sequences, surpassing them across all evaluation metrics.

In summary, we make the following contributions:
\begin{itemize}[left=0pt]

\item We show that optimizing object-centric knowledge of generative models enhances their ability to generate complex motions and interactions, suggesting a promising direction for improving video generation.

\item We introduce \modelname, a three-stage pipeline that significantly improves the quality of pre-trained video generation models by explicitly optimizing parameterized 3D object-centric motion knowledge extracted from generated videos.

\item We propose \motionmodelname, a lightweight and robust parameterized motion prior model that effectively refines motion information in generated videos.

\end{itemize}

\section{Related Work}
\label{sec:relatedWork}
\noindent \textbf{Video Generation.}
With the success of diffusion models in image generation~\citep{rombach2022high, esser2024scaling, liu2024playground, betker2023improving}, driven by advancements in both generative modeling strategies~\citep{ho2020denoising, song2020score, lipman2022flow, liu2022flow, liu2024flowing} and model architectures~\citep{bao2023all, peebles2023scalable, liu2024alleviating, ma2024sit}, video generation~\citep{ho2022video, singer2022make, wang2024lavie, yang2023probabilistic, zhang2024show, zhou2022magicvideo, bar2024lumiere, polyak2024moviegencastmedia, brooks2024video}  has recently attracted significant attention.
Parallel to text-to-video (T2V) generation, image-to-video (I2V) methods~\citep{babaeizadeh2017stochastic, li2018flow, xiong2018learning, pan2019video, zhang2020dtvnet} generate videos from a single starting frame.
However, existing methods still struggle to handle complex motions and interactions, and often fail to maintain physical plausibility.
To overcome these challenges, recent approaches incorporate additional conditions to enhance motion control in video generation.
Common conditional inputs include text descriptions~\citep{hu2022make, girdhar2023emu, chen2023seine, ren2024consisti2v, zeng2024make}, which can further guide motion modeling.
For example, MAGE~\citep{hu2022make} introduces a spatially aligned motion anchor to blend motion cues from text, and SEINE~\citep{chen2023seine} uses a random-mask video diffusion model to create transitions guided by textual descriptions. 
Another popular condition is optical flow, where models~\citep{mahapatra2022controllable, ni2023conditional, shi2024motion} estimate rough flow from user-provided arrows or text to guide complex motion generation.
In contrast, \modelname leverages implicit motion features already embedded in the generated video through 3D parameterized object representations.
This allows it to directly extract, optimize, and reinforce accurate and reliable motion features from the generated video itself, resulting in precise motion sequences that enhance coherence and fidelity.

\noindent \textbf{Human Image Animation.}
Human image animation focuses on transferring motion from a source human to a target human by \textit{using ground-truth posture sequences}, which can be represented as flow~\citep{wang2004image}, keypoints~\citep{hu2024animate,tan2024animate}, or human part masks~\citep{xu2024magicanimate}. 
Extensive efforts have gone into extracting improved motion features. 
For example, MagicAnimate~\citep{xu2024magicanimate} leverages an off-the-shelf ControlNet~\citep{zhang2023adding} to obtain motion conditions, Hu \etal~\citep{hu2024animate} introduce a Pose Guider network to align pose images with noise latents, and Animate-X~\citep{tan2024animate} utilizes both implicit and explicit pose indicators to generate motion and pose features. 
Such strong guidance enables high-quality video generation in human image animation, as each posture sequence directly dictates the synthesis of corresponding frames. 
However, \textit{in general video generation, the ground-truth dense guidance is typically unavailable}, and there is usually no reference video for extracting a motion sequence.
To overcome this limitation, \modelname introduces a three-stage process: it first extracts an implicit, rough motion sequence from the generated video, then refines it using the proposed \motionmodelname, and finally leverages the refined motion to guide video regeneration.
This approach provides effective guidance for video generation, significantly improving video quality.

\section{Preliminary}
\label{sec:preliminaries}
\noindent\textbf{Latent Diffusion Model.}
Diffusion models~\citep{ho2020denoising} generate data through a denoising process that learns a probabilistic transformation. Latent diffusion models~\citep{rombach2022high} move this process from pixel space to the latent space of a Variational Autoencoder~\citep{kingma2013auto}. Specifically, we consider the latent representation $z_0$ of the input data. In the forward diffusion process, Gaussian noise is incrementally added to $z_0$:
\begin{equation}
    q(z_t | z_{t-1}) = \mathcal{N}\left(z_t; \sqrt{1 - \gamma_t} z_{t-1}, \gamma_t \mathbf{I}\right),
\end{equation}
where $z_t$ represents the noisy latent representation at time step $t$, and $\gamma_t$ is a predefined noise schedule with $t \in (0, 1)$. As $t$ increases, the cumulative noise applied to $z_0$ intensifies, gradually transforming $z_t$ closer to pure Gaussian noise. We express the transformation from  $z_0$ to $z_t$ directly as: $z_t = \sqrt{\bar{\alpha}_t} z_0 + \sqrt{1 - \bar{\alpha}_t} \, \epsilon$, 
where $\bar{\alpha}_t = \prod_{i=1}^t (1 - \gamma_i)$ and $\epsilon \sim \mathcal{N}(0, \mathbf{I})$. The latent diffusion model, parameterized by $\Theta$, learns to reverse this noising process by taking $z_t$ as input and reconstructing the clean data with the objective: $\mathcal{L} = \left\| \epsilon - \epsilon_\Theta(z_t, t, c) \right\|_2^2$,
where $c$ is the condition to guide the denoising process. 
Once the latent space is reconstructed, it is decoded via the VAE decoder.

\noindent\textbf{Video Latent Diffusion Model.}
We use SVD~\citep{blattmann2023stable} as our base video diffusion model, which extends Stable Diffusion 2.1~\citep{rombach2022high} to video. The main architectural difference from image diffusion models is that SVD incorporates a temporal UNet~\citep{ronneberger2015u} by adding temporal convolution and (cross-) attention~\citep{vaswani2017attention} layers after each corresponding spatial layer.

\noindent\textbf{3D Human and Animal Mesh Recovery.}
We utilize the SMPL-X~\citep{pavlakos2019expressive} and SMAL~\citep{zuffi20173d} parametric models to represent humans and animals, respectively.
These models parameterize 3D meshes with pose parameters $\theta$ and shape parameters $\beta$. 
Additionally, SMPL-X model includes expression parameters $\psi$ to capture facial expressions through blend shapes.
Given these parameters, SMPL-X model is a differentiable function that outputs a posed 3D human mesh $\mathcal{M}_{SMPL-X}(\theta_h, \beta_h, \psi_h) \in \mathbb{R}^{10475 \times 3}$, where pose $\theta_h \in \mathbb{R}^{165}$, shape $\beta_h \in \mathbb{R}^{10}$, and expression $\psi_h \in \mathbb{R}^{10}$. 
Similarly, SMAL model represents a posed 3D animal mesh with $\mathcal{M}_{SMAL}(\theta_a, \beta_a) \in \mathbb{R}^{3889 \times 3}$, where pose $\theta_a \in \mathbb{R}^{105}$ and shape $\beta_a \in \mathbb{R}^{41}$. 
In our work, we recover 3D human and animal meshes by fitting the SMPL-X and SMAL models to both our data and the generated videos. This produces 3D mesh reconstructions for all humans and animals. Because these meshes are computer-graphics models with predefined body-part annotations at every vertex, we can obtain accurate part labels directly. The 3D meshes also allow us to compute motion strength by measuring movement speed in 3D space, which is more reliable than relying on 2D pixel motion alone.

\noindent\textbf{2.5D Parameterized Object Representation.}
Unlike humans and animals, there is no straightforward way to parameterize general objects in 3D space. 
Here, we represent objects in 2.5D by combining 2D bounding boxes~\citep{varghese2024yolov8}, segmentation masks~\citep{peng2020deep, wu2024general}, and estimated depth~\citep{depth_anything_v2}.

\section{Method}
\label{sec:method}

\begin{figure*}
    \centering
    \includegraphics[width=\linewidth]{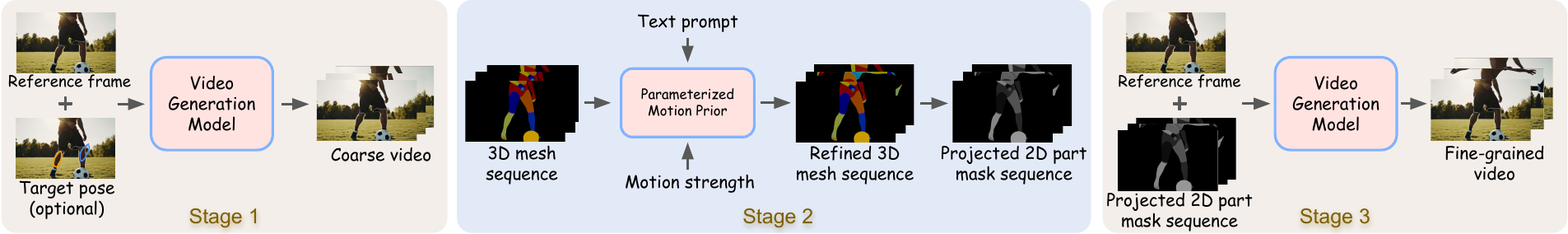}
    \caption{\textbf{Method overview.}
    Given the video generation model, \modelname operates in three stages. Stage 1: A coarse video is generated based on the provided conditions (\eg, target pose, marked in \textcolor{blue}{blue}, indicating the rough position of the \textcolor[HTML]{ffca33}{yellow} part in the last frame). Stage 2: 3D features from the generated coarse video are extracted and optimized using the proposed \motionmodelname.
    Stage 3: The optimized 3D sequences are used to regenerate the video with enhanced motion consistency. Best viewed when zoomed in.
    }
    \label{fig:pipeline}
\end{figure*}

\modelname requires extending a pre-trained video diffusion model to accept additional motion conditions as input. 
In Sec.~\ref{subsec:svd}, we describe how to adapt SVD into a motion-conditioned video generation model with minimal modifications. 
In Sec.~\ref{subsec:revision}, we introduce \modelname, a three-stage video generation pipeline built upon the extended SVD, incorporating a Parameterized Motion Prior model (\motionmodelname) to provide accurate motion sequences as conditioning inputs.
An illustration is provided in Fig.~\ref{fig:pipeline}.

\subsection{Motion-Conditioned Video Generation}
\label{subsec:svd}
\begin{wrapfigure}{r}{0.42\textwidth}
    \vspace{-5mm}
    \includegraphics[width=0.42\columnwidth]{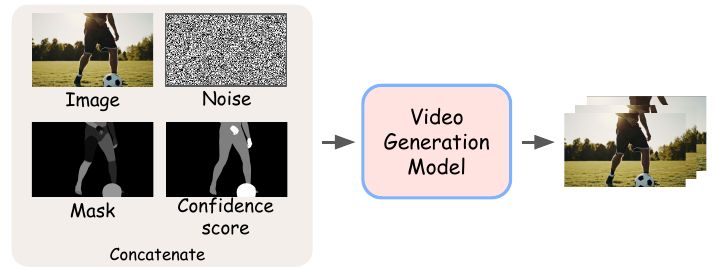}
    \caption{\textbf{Motion-conditioned video generation.}
    We enable motion-conditioned generation by introducing two extra conditioning channels: (1) part segmentation mask derived from the 3D motion sequence, and (2) its corresponding confidence map.
    }
    \vspace{-3mm}
    \label{fig:pipeline_vidgen}
\end{wrapfigure}

Since SVD does not natively support motion-conditioned video generation, we extend its design to enable this capability, with a focus on simplicity to preserve its original generation quality and minimize deviations from user-provided inputs.
Concretely, we begin with a pre-trained SVD and fine-tune it within a carefully structured strategy. 
We concatenate two additional conditioning channels to the original condition: one for a part-level segmentation mask derived from the 3D motion sequence, and another for a confidence map indicating the reliability of the part mask, as illustrated in Fig.~\ref{fig:pipeline_vidgen}. 
We also design a fine-tuning pipeline that integrates three scenarios with varying levels of part mask guidance, allowing the model to flexibly handle diverse inputs.
We detail those three scenarios below.

First, when the full motion sequence is provided (40\% of training examples), the part-level mask is generated by merging all 2D part segmentation masks projected from 3D parametric mesh models.
Since the motion sequence provides dense and precise control over video generation, we assign a confidence score of 1 to the corresponding confidence map.
Our experiments confirm that these 3D-projected masks are more robust than existing part segmentation models.

Second, when only the target pose is provided (30\% of training examples), we convert the projected part segmentation masks into polygons. 
This aligns with users' inference input, where they provide simple sketches (\eg, circles or ovals) to indicate the final positions of specific targets or parts (\eg, a hand or arm). These user-friendly sketches are then converted into polygonal masks, similar to the part segmentation mask polygons used during training.
Since polygon conversion introduces unavoidable errors, we assign a confidence score of 0.5 in this case.

Last, to preserve SVD's ability to generate videos without motion conditioning, the remaining 30\% of training examples provide an empty part mask, with a corresponding confidence score of 0.

Note that all three settings use the same model architecture, with minimal modifications limited to the first convolutional block of SVD.
This design enables fine-tuning only the initial convolutional block and the temporal layers, avoiding the need to train SVD from scratch.
As a result, the extended SVD can generate videos conditioned on various types of motion inputs, while still retaining its ability to generate videos from just text and the first frame.

\subsection{Proposed Method: ReVision}
\label{subsec:revision}

\textbf{Overview.} As shown in Fig.~\ref{fig:pipeline}, \modelname consists of three stages. 
In stage one (S1), we generate a coarse video based on the provided conditions. 
In stage two (S2), we extract both 2D and 3D features from the coarse video and refine the 3D motion sequences through the proposed \motionmodelname. 
In stage three (S3), we use the refined 3D motion sequences as strong conditioning, guiding the video generation model to regenerate the video, resulting in significantly higher-quality output even for complex motions and interactions. 

\noindent \textbf{S1: Coarse Video Generation.}
Given the first frame and an \textit{optional} user-specified target motion in the final frame, we use the fine-tuned SVD model to generate the video. 
Since the generation relies only on the target motion in the final frame or an empty motion, rather than a complete motion sequence, the resulting video often exhibits poor motion quality, leaving room for refinement. 
Therefore, we refer to this stage as coarse video generation.

Although we utilize only 2D and 3D motion features from the coarse video generated in Stage 1, this phase remains foundational, as it is critical for capturing rich motion patterns, intricate object interactions, and authentic camera movements in complex real-world settings.
Video generation is inherently a multifaceted task. Beyond producing realistic object appearances, it also requires an understanding of dynamic motion, scene context, coherent camera trajectories, and the diverse interplay of these elements. While training a motion generation model directly for this task is theoretically feasible, it proves challenging in practice. Current state-of-the-art models~\citep{guo2024momask, zhang2024motiondiffuse, zhang2023remodiffuse} are constrained by the lack of large datasets, limiting their capacity to model only simplistic motions, such as human-like activities like running or dancing. Consequently, they struggle to generalize to more complex motions, diverse object interactions, and fail to generate motions that align with realistic camera dynamics and scene context.

Stage 1 mitigates these limitations by directly generating videos and extracting detailed motion patterns, camera trajectories, scene transitions, and meaningful interactions. By leveraging extensive motion priors learned from billions of videos, it constructs a comprehensive sketch of motion and scene structure. This sketch serves as a critical foundation, providing the diversity and realism necessary to produce lifelike, engaging videos. The subsequent stages build on this: Stage 2 refines the motion further, while Stage 3 focuses on generating the video’s visual appearance based on the refined motion feature.

Notice that, because only rough motion features without detailed visual information are needed from the coarse video, \textbf{we can significantly reduce computational overhead}. For instance, our experiments show that the compute time for Stage 1 can be reduced from 36 seconds to 8 seconds by generating the coarse video at a lower resolution (1/4 of the original), with fewer frames (1/2 of the original), and fewer denoising steps (32 vs. 50), while still preserving comparable final video quality. This optimization makes the overall generation process more efficient and cost-effective (See Tab.~\ref{tab:speed}).

\noindent \textbf{S2: Object-Centric 3D Optimization.}
After generating coarse videos, we parameterize the 3D information in the scenes for further optimization.
For \textit{humans} and \textit{animals}, we employ well-established 3D parametric mesh models~\citep{loper2023smpl, rueegg2023bite, Zuffi:CVPR:2024}.
For \textit{general objects}, where no unified 3D representation or well-established modeling approach exists, we construct a parameterized representation by combining 2D bounding boxes~\citep{varghese2024yolov8}, segmentation masks~\citep{yu2023convolutions, peng2020deep}, and estimated depth~\citep{depth_anything_v2}.
Specifically, given the detected bounding box and segmentation mask, we extract a contour from the mask and approximate it with 16 vertices. 
We then combine these with 4 bounding box corners and the box center, yielding a total of 21 key 2D points. 
These points are lifted into 3D space using the estimated depth, resulting in a compact point-based representation for each object, denoted as $p_o\in\mathbb{R}^{21\times 3}$.

However, due to the poor motion quality and inconsistencies in the coarse video generated in S1, the 3D parameters extracted also suffer from instability and inconsistencies.
To address this, we propose a Parameterized Motion Prior model (\motionmodelname) to optimize the 3D motion sequence, based on the text information and the motion strength.

\motionmodelname first extracts text embeddings from the text description using a pre-trained CLIP encoder~\citep{radford2021learning}.
Motion strength is computed from the differences in parametric 3D model parameters between adjacent frames, providing a measure of motion speed.
Since the 3D motion sequences are already parameterized as 3D vectors, \motionmodelname employs a series of transformer blocks to iteratively refine the motion sequence based on these conditioning inputs.
Within each block, motion features undergo self-attention, followed by cross-attention with the conditioning inputs (text embeddings and motion strength) and a feedforward network to generate the final output. 
Finally, the optimized 3D parameterized motion sequences are converted into 3D mesh sequences and projected into 2D as part segmentation masks and confidence maps, providing more accurate motion guidance.
Architectural details are provided in Sec~\ref{sec:supp:PPPM} of the Appendix, and the effectiveness of \motionmodelname is demonstrated in Sec.~\ref{subsec:exp:PPPM}.

To train \motionmodelname, we introduce small perturbations to the ground-truth motion sequences from the annotated Panda-70M subset.
Three types of perturbations are randomly applied: (1) adding small random noise to the motion sequence, (2) shuffling the internal order of the sequence, and (3) dropping a small segment while repeating the remaining segments to maintain the original length.
Through this process, \motionmodelname learns to denoise perturbations, improving its ability to recover smooth and robust motion sequences.

\noindent \textbf{S3: Fine-grained Video Generation.}
In the final stage, we regenerate the video using the same SVD model but with the improved motion sequence as additional conditioning. 
Unlike the coarse generation in stage one, which uses only the target pose in the last frame or no motion information, we now utilize the full motion sequence as part masks optimized in 3D space.
With this stronger conditioning, the final output exhibits significantly improved motion consistency compared to the coarse video, as illustrated in Fig.~\ref{fig:abl_stage1-3}.

\section{Experimental Results}
\label{sec:experiments}
\begin{figure*}[ht]
    \centering
    \includegraphics[width=\linewidth]{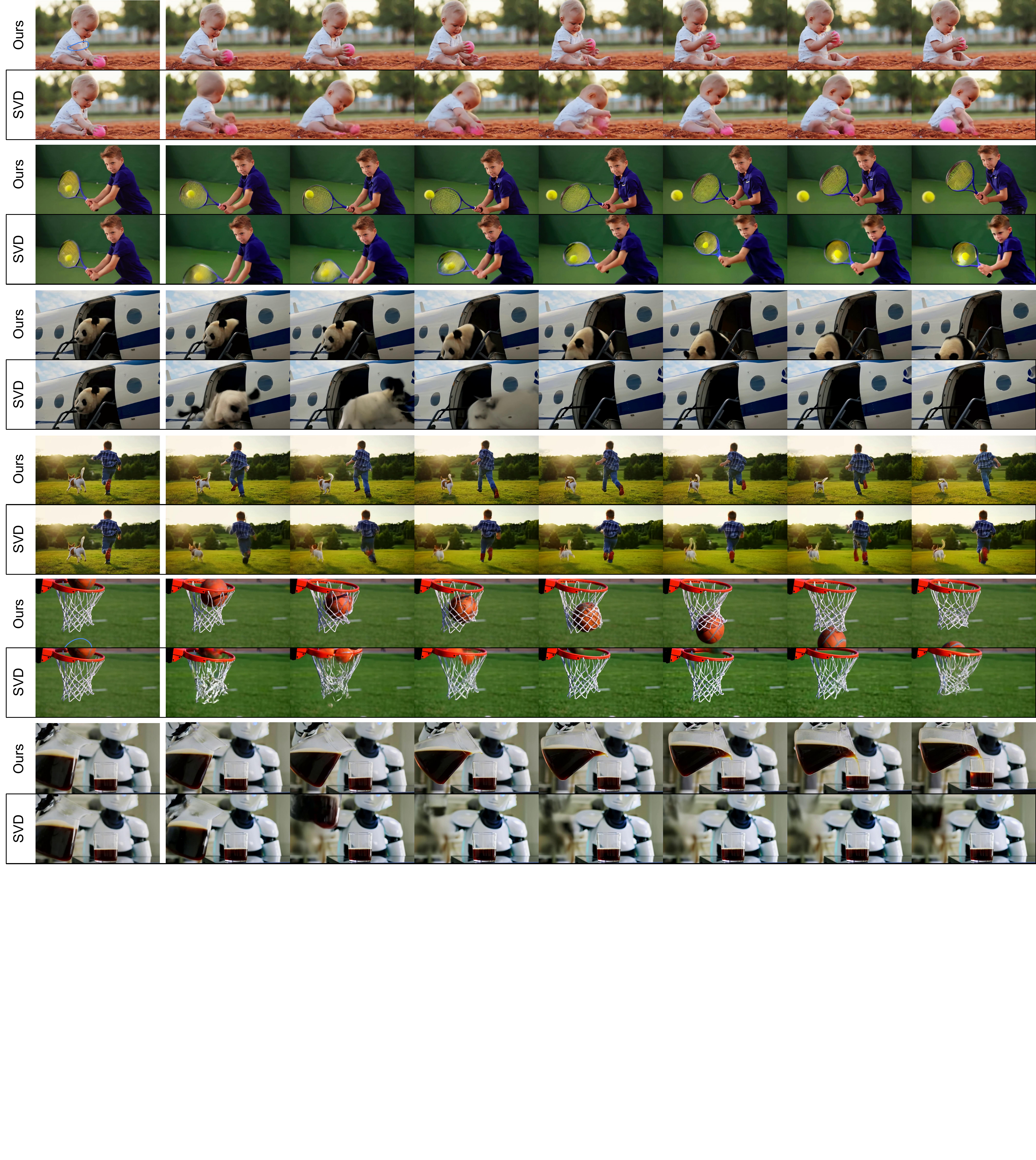}
    \vspace{-48mm}
    \caption{\textbf{Qualitative comparisons.}
    \modelname generates high-quality videos with complex motions and interactions of \textit{humans, animals, and general objects}.
    Zoom in for better details. Please find the side-by-side video comparisons in the supplementary video. Reference frames are in the first column.
    }
    \label{fig:vis}
\end{figure*}

We first compare our method with SVD~\citep{blattmann2023stable} and HunyuanVideo~\citep{kong2024hunyuanvideo} in Sec.~\ref{subsec:exp:I2V}, highlighting how it enhances SVD to support more controllable and complex motion generation while maintaining efficiency, effectively handling occlusions, and enabling long video generation.
Next, in Sec.~\ref{subsec:exp:CA}, we compare our model with Human Image Animation methods, demonstrating its ability to generate complex motions.
We then evaluate the effectiveness of the proposed Parameterized Motion Prior in Sec.~\ref{subsec:exp:PPPM}.
Due to space limitations, additional details and results, including ablations on parametric 3D mesh, text prompt, and motion strength, are provided in Sec.~\ref{sec:supp:abl} of the Appendix.

\subsection{Image-to-Video Generation}
\label{subsec:exp:I2V}
\noindent\textbf{Dataset.}
Both the motion-conditioned video generation model and the Parameterized Motion Prior model (\motionmodelname) need to be fine-tuned (trained) on a small yet high-quality video dataset with object-centric annotations.
Existing large-scale video datasets~\citep{bain2021frozen,chen2024panda} mainly provide text-image pairs without detailed object-centric annotations.
To address this limitation, we use a suite of off-the-shelf models across various tasks to generate 2D and 3D object-centric annotations. 
We annotate a total of 20K videos from the Panda-70M~\citep{chen2024panda} dataset.
For each video, we provide frame-wise 2D bounding boxes, semantic masks, depth estimation maps, and 3D parametric mesh reconstructions for detected humans and animals.
The details are outlined in Sec.~\ref{sec:dataset} in the Appendix.

\noindent\textbf{Experimental Setup.}
For most experiments, we use SVD~\citep{blattmann2023stable} as the base video generation model and modify it by introducing two additional channels for conditional generation.
We fine-tune SVD on our annotated dataset for 300K iterations with a batch size of 64 and a constant learning rate of $2 \times 10^{-5}$.
During training, we randomly sample 16-frame video clips with a stride of 4 at a resolution of $1024 \times 576$. 
To enable various control, we incorporate different conditioning strategies: $40\%$ of video clips contain accurate part masks for each frame, $30\%$ contain a polygon mask for random parts in the final frame, and the remaining clips have no additional conditioning.

\begin{figure}[t]
\centering

\begin{minipage}[t]{0.48\textwidth}
\centering
\begin{tikzpicture}
\begin{axis}[
    ybar,
    bar width=14pt,
    ymajorgrids=true,
    grid style={draw=gray!30},
    width=1.1\linewidth,
    height=0.55\linewidth,
    enlarge x limits=0.2,
    ylabel style={font=\scriptsize, yshift=-15pt},
    ylabel={User preference},
    ymin=0, ymax=100,
    ytick distance=20,
    minor y tick num=3,
    tick label style={font=\scriptsize},
    symbolic x coords={Motion Consistency, Amount of Motion, Motion Reality},
    xtick=data,
    legend style={
        at={(0.5,1.02)},
        anchor=south,
        legend columns=-1,
        font=\scriptsize 
    },
]
\addplot[fill=NavyBlue] coordinates {
  (Motion Consistency,62.31)
  (Amount of Motion,93.88)
  (Motion Reality,82.99)
};
\addplot[fill=orange] coordinates {
  (Motion Consistency,37.69)
  (Amount of Motion,6.12)
  (Motion Reality,17.01)
};
\legend{\modelname-SVD (Ours)$\quad$, SVD}
\end{axis}
\end{tikzpicture}
\end{minipage}
\hfill
\begin{minipage}[t]{0.48\textwidth}
\centering
\begin{tikzpicture}
\begin{axis}[
    ybar,
    bar width=14pt,
    ymajorgrids=true,
    grid style={draw=gray!30},
    width=1.1\linewidth,
    height=0.55\linewidth,
    enlarge x limits=0.2,
    ylabel style={font=\scriptsize, yshift=-15pt},
    ylabel={User preference},
    ymin=0, ymax=80,
    ytick distance=20,
    minor y tick num=3,
    tick label style={font=\scriptsize},
    symbolic x coords={Motion Consistency, Amount of Motion, Motion Reality},
    xtick=data,
    legend style={
        at={(0.5,1.02)},
        anchor=south,
        legend columns=2,
        font=\scriptsize 
    },
]
\addplot[fill=NavyBlue] coordinates {
  (Motion Consistency,64.13)
  (Amount of Motion,54.12)
  (Motion Reality,62.56)
};
\addplot[fill=orange] coordinates {
  (Motion Consistency,35.87)
  (Amount of Motion,45.88)
  (Motion Reality,37.44)
};
\legend{\modelname-SVD (Ours)$\quad$, HunyuanVideo}
\end{axis}
\end{tikzpicture}
\end{minipage}

\vspace{-2mm}
\caption{\textbf{User preference comparisons.} Our model enhances the motion generation capability of the pre-trained SVD. It even surpasses HunyuanVideo, a SOTA model with 13B parameters. These results highlight the effectiveness of our model in generating complex motions and interactions.}
\label{fig:user_comparison}
\end{figure}

\begin{table}[]
\centering
\caption{\textbf{Quantitative comparisons on VBench++.} We achieve a significantly higher Dynamic Degree while maintaining similar performance across all metrics of consistency, smoothness, and quality.}
\resizebox{\columnwidth}{!}{
\begin{tabular}{l|c|cccccccc}
 & {\bf Model} & {\bf I2V} & {\bf I2V} & {\bf Subject} & {\bf Background}  & {\bf Motion}  & \cellcolor{yellow!25} {\bf Dynamic} & {\bf Imaging }  \\ 
 & {\bf Type}  &  {\bf Subject} & {\bf Background} &  {\bf Consistency} &  {\bf Consistency} &  {\bf Smoothness} &  \cellcolor{yellow!25}  {\bf Degree} &  {\bf Quality}  \\ 
\midrule
Step-Video-TI2V~\citep{huang2025step}    & TI2V & 98.63\% & 98.63\% & 96.02\% & 97.06\% & 99.24\% & \cellcolor{yellow!25} 48.78\% & 70.44\% \\
DynamiCrafter-1024~\citep{xing2024dynamicrafter} & TI2V & 98.17\% & 98.60\% & 95.69\% & 97.38\% & 97.38\% & \cellcolor{yellow!25} 47.40\% & 69.34\% \\
\midrule
Gen-4-I2V~\citep{gen4} & I2V & 97.84\% & 97.46\% & 93.23\% & 96.79\% & 98.99\% & \cellcolor{yellow!25}  55.20\% & 70.41\% \\
Magi-1~\citep{teng2025magi} & I2V & 98.39\% & 99.00\% & 93.96\% & 96.74\% & 98.68\% & \cellcolor{yellow!25} 68.21\% & 69.71\% \\
HunyuanVideo-I2V~\citep{kong2024hunyuanvideo} & I2V & 98.53\% & 97.37\% & 95.26\% & 96.70\% & 99.23\% & \cellcolor{yellow!25}  22.20\% & 70.1\% \\
\midrule
Wan2.1-I2V-14B-720P~\citep{wan2025} & I2V & 96.95\% & 96.44\% & 94.86\% & 97.07\% & {\bf 97.90\%} & \cellcolor{yellow!25}  51.38\% & 70.44\%  \\
{\bf ReVision-Wan2.1  (Ours)}  & I2V & {\bf 98.10\%} & {\bf 97.10\%} & {\bf 97.06\%} & {\bf 97.89\%} & 97.74\% & \cellcolor{yellow!25} {\bf 73.67\%} & {\bf 72.86\%} \\
\midrule
SVD-XT-1.1~\citep{blattmann2023stable} & I2V & 97.51\% & 97.62\% & 95.42\% & 96.77\% & 98.12\% & \cellcolor{yellow!25}  43.17\%  & 70.23\% \\
{\bf ReVision-SVD (Ours)}  & I2V & {\bf 97.94\%} & {\bf 98.06\%} & {\bf 96.13\%} & {\bf 97.89\%}   & {\bf 98.88\%}   & \cellcolor{yellow!25}  {\bf 83.15\%}  & {\bf 71.48\%} \\

\end{tabular}}
\label{tab:vbench}
\vspace{-3mm}
\end{table}

\noindent\textbf{Benchmark on General Video Generation.} To better evaluate our model on general video generation, we used VBench++~\citep{huang2024vbench}, which provides a comprehensive, detailed, multi-dimensional assessment of general video generation quality. 
As our model is designed for image-to-video generation (I2V), we primarily compared it against SVD-XT-1.1~\citep{blattmann2023stable}. 
Moreover, since our model is backbone-agnostic, we integrate our method with a stronger and more recent video diffusion backbone, Wan2.1-I2V-14B-720P~\citep{wan2025}. We train \modelname-Wan2.1 on the same dataset as \modelname-SVD using the same settings.
Results are provided in Tab.~\ref{tab:vbench}. 
Our models consistently outperforms the base models on nearly all metrics, particularly in dynamic degree (83.15\% vs. 43.17\%, 73.67\% vs. 51.38\%) and various consistency and smoothness measures.

In addition, we compared our method with recent state-of-the-art models for text+image-to-video (TI2V) generation in Tab.~\ref{tab:vbench}. 
We observe that, although these TI2V models use text descriptions to specify motion, the intensity and quality of the resulting movements are not fully controllable. 
In contrast, our method achieves substantially better performance in generating complex motions, while maintaining high consistency, temporal smoothness, and overall visual quality.

\begin{table}[t]
\centering
\caption{\textbf{Comparisons with motion-conditioned video generation methods on DAVIS.} \modelname achieves more accurate motion transfer while preserving object appearance and scene details, resulting in enhanced temporal coherence and reduced visual artifacts.
}
\label{tab:motion_conditioned_comparison}
\resizebox{0.9\linewidth}{!}{%
\begin{tabular}{l|cccccc}
& {\bf CoTracker} & {\bf Optical Flow} & {\bf Pixel} & {\bf Subject}  & {\bf Background}  & {\bf Motion}  \\ 
&  {\bf mIoU $\uparrow$} & {\bf Error $\downarrow$} &  {\bf MSE $\downarrow$} &  {\bf Consistency $\uparrow$} &  {\bf Consistency $\uparrow$} &  {\bf Smooth $\uparrow$}  \\ 
\midrule
MotionClone~\citep{ling2024motionclone} & 0.72 & 0.42 & 0.068 & 0.75 & 0.85 & 0.92 \\
ImageConductor~\citep{li2025image} & 0.66 & 0.64 & 0.072 & 0.77 & 0.88 & 0.93 \\
Go-with-the-Flow~\citep{burgert2025go} & 0.74 & 0.36 & 0.053 & 0.88 & 0.92 & 0.98 \\
\midrule
{\bf \modelname-SVD (Ours)}  & {\bf 0.80} & {\bf 0.33} & {\bf 0.046} & {\bf 0.96} & {\bf 0.97} & {\bf 0.99} \\
\end{tabular}}
\end{table}
\textbf{Comparisons with Motion-conditioned Video Generation.}
Following Go-with-the-Flow~\citep{burgert2025go}, we also consider the motion-transfer image-to-video task on DAVIS~\citep{pont20172017} and report the results in Tab.~\ref{tab:motion_conditioned_comparison}. We observe that our model consistently outperforms all baselines across all metrics, clearly demonstrating the effectiveness of the proposed method. In particular, our approach yields more accurate motion transfer while better preserving object appearance and scene details, leading to improved temporal coherence and fewer visual artifacts.

In addition, comparing text-conditioned and motion-conditioned video generation shows that explicit motion conditioning plays a dominant role in controlling dynamics, while text conditioning provides only coarse guidance. Although text descriptions effectively convey high-level semantic intent, they struggle to specify fine-grained temporal details and precise motion patterns. As a result, \emph{motion conditioning enables more direct and reliable control over dynamic behaviors that are difficult to express through natural language.}

\noindent\textbf{User Study.}
To better compare our model with the baseline SVD and HunyuanVideo, we conduct user studies to assess user preferences.
Specifically, we generate $500$ text descriptions of humans and animals engaged in daily activities using GPT-4o~\citep{hurst2024gpt}.
For the comparison with SVD, we use GPT-4o to generate five $16:9$ images for each prompt, which are resized to $1024 \times 576$ as input.
For the comparison with HunyuanVideo, we first use their released model to generate five videos at a resolution of $1280 \times 720$ for each prompt, then extract the first frame of each video as the input image for our model to generate the corresponding video.
No target pose is provided for any model.
For each image, we generate one video per model using the same random seed ($42$), resulting in a total of $5,000$ video pairs.
Each video pair is evaluated by three randomly selected users on Amazon MTurk, leading to a total of $15,000$ comparisons.
Users are shown two videos side by side, generated by different models, with the order randomized.
They are instructed to assess the videos based on Motion Consistency, Amount of Motion, and Motion Realism.
The results are reported in Fig.~\ref{fig:user_comparison}.
Our model significantly enhances the motion generation capabilities of SVD, producing videos with superior motion quantity, consistency, and realism. 
Furthermore, it even surpasses HunyuanVideo, a state-of-the-art video generation model with 13B parameters, in terms of motion quality.
These results highlight the effectiveness of our model in generating complex motions and interactions.

\noindent\textbf{Qualitative Comparisons.} 
We provide samples of generated videos in Fig.~\ref{fig:vis}.
Our \modelname produces realistic movements that closely follow user instructions.
It also generates high-quality videos that involves complex motions and interactions, such as running with dogs, picking up a ball, and hitting a tennis ball.
More visualizations are available in the supplementary videos.

\begin{wraptable}{r}{0.4\columnwidth}
\vspace{-4.5mm}
  \centering
  \caption{{\small \textbf{Inference speed.} Average time to generate a 32-frame video. Our \modelname-SVD matches SVD in speed (8.4x faster than HunyuanVideo) while surpassing HunyuanVideo in generating complex motions and interactions.}}
  \label{tab:speed}
  \resizebox{\linewidth}{!}{%
    \begin{tabular}{l|ccc}
       & {\bf SVD} & {\bf \modelname-SVD} & {\bf HunyuanVideo} \\
      \midrule
      Time (s) & 36 & 49 (8 + 5 + 36) & 411 \\
    \end{tabular}%
  }
  \vspace{-3mm}
\end{wraptable}

\noindent\textbf{Inference Speed.} 
We compare the inference speed of our model against two baselines in Tab.~\ref{tab:speed}.
Despite the three-step pipeline, the coarse video generation (S1) takes only 8 seconds after our optimization, and the additional 3D detection and refinement modules (S2) add 5 seconds to the inference time on a single A100. Together, these two stages are significantly faster than the original SVD, which requires 36 seconds.
More importantly, with just 1.5B parameters and a runtime of 49 seconds, our model generates high-quality videos with complex motions -- comparable to or even surpassing state-of-the-art models like HunyuanVideo (see Fig.~\ref{fig:vis} and ~\ref{fig:user_comparison}), which uses over 13B parameters and requires an average of 411 seconds.

\begin{figure}
    \centering
    \includegraphics[width=\linewidth]{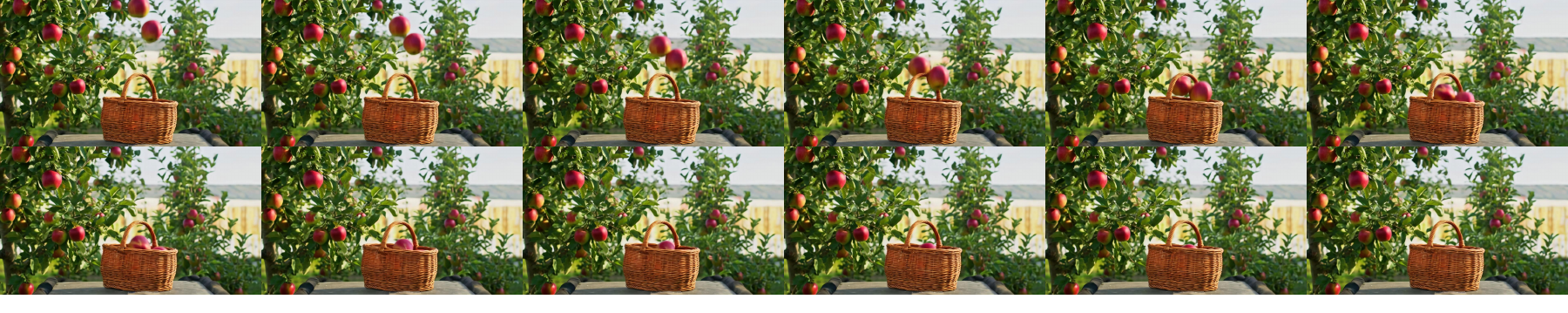}
    \vspace{-7mm}
    \caption{\textbf{Handling occlusion.}
    As illustrated by the two apples falling into the basket, \modelname handles occlusions by lifting and optimizing motion in 3D space, which allows explicit reasoning about object spatial relationships, effectively resolving occlusions that are ambiguous in 2D.}
    \label{fig:occ}
\end{figure}
\noindent{\textbf{Handling Occlusion.}} 
Occlusion becomes a significant challenge when generating videos with multiple objects and large motions. 
However, by lifting everything into 3D, occlusion is naturally resolved: Since we estimate depth, all objects are fully represented with their spatial positions, allowing us to reason about their relative locations in 3D.
And after optimizing the motion in 3D, we project it back to 2D using the depth information, which restores accurate occlusion relationships in the camera coordinate.
An example is shown in Fig.~\ref{fig:occ}, where two apples are generated dropping into a basket. 
Our model effectively captures spatial relationships, producing realistic videos in which the apples fall \emph{into} the basket with appropriate occlusion.

\begin{wraptable}{r}{0.4\columnwidth}
\vspace{-4.5mm}
  \centering
  \caption{{\small \textbf{User preference comparisons for occlusion and interaction handling.}}}
  \label{tab:user_occ}
  \resizebox{\linewidth}{!}{%
    \begin{tabular}{l|ccc}
       & {\bf \modelname-SVD (Ours)} & {\bf SVD} \\
      \midrule
      Preference &  97.63\% & 2.37\% 
      \vspace{3mm} \\
       & {\bf \modelname-SVD (Ours)} & {\bf HunyuanVideo} \\
      \midrule
      Preference &  63.99\% & 36.01\% \\
    \end{tabular}%
  }
  \vspace{-3mm}
\end{wraptable}

To further evaluate our model's ability to handle occlusion, we conducted an additional user study using the 5,000 video pairs generated for the main experiment.
Users on Amazon MTurk were asked to assess the quality of occlusion handling and object interactions for each video pair. 
If no object interaction was observed, users were instructed to select "no interaction/occlusion."
Similarly, each video pair was evaluated independently by three randomly selected users to ensure reliability.
The study yielded 15,000 evaluations, including 10,207 valid comparisons and 4,793 responses marked as "no interaction/occlusion". 
The results of these 10,207 valid comparisons are summarized in Tab.~\ref{tab:user_occ}, highlighting that our model consistently outperforms both SVD and HunyuanVideo across a diverse range of occlusion scenarios.

\begin{figure}
    \centering
    \includegraphics[width=\linewidth]{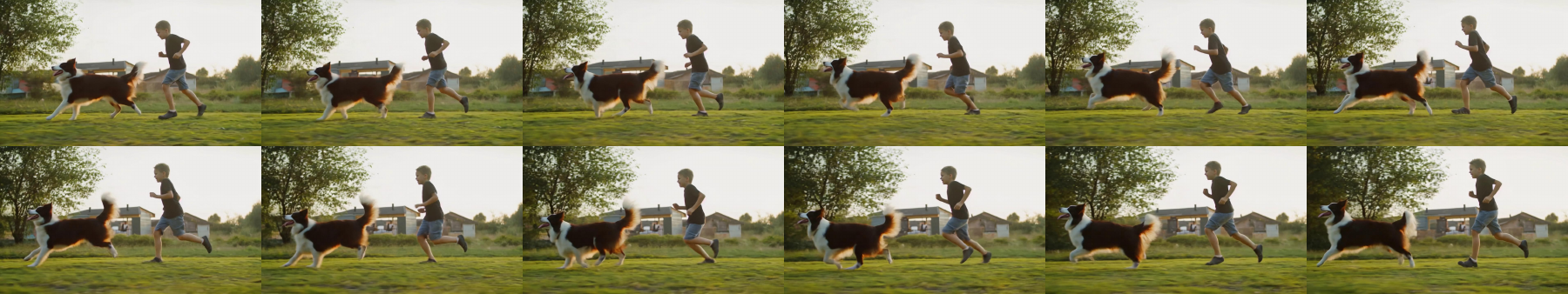}
    \vspace{-3mm}
    \caption{\textbf{Long video generation.} Our \motionmodelname extends a 32-frame 3D motion to 128 frames through interpolation (32 $\rightarrow$ 64), extrapolation (64 $\rightarrow$ 128), and refinement, enabling complex, large-scale motion generation over long video sequences. See supplementary videos for details.}
    \label{fig:long}
\end{figure}
\noindent{\textbf{Long Video Generation.}} 
Another advantage of our model is its ability to generate complex and large-scale motions over long video sequences (Fig.~\ref{fig:long}).
\motionmodelname optimizes motion in a 3D parameterized space, enabling smooth and realistic interpolation and extrapolation to arbitrary lengths. The resulting long 3D motion sequences are then used to generate multiple overlapping video clips, which are stitched together to form extended videos with consistent motion. 
More specifically, we resample in motion-parameter space (not pixel space) through interpolation and extrapolation. Then, for long-video synthesis we generate overlapping 32-frame clips with a sliding window (stride 24, overlap 8) and stitch overlaps with a ramped blend in the overlapping frames. Motion conditions are aligned in the overlap and all windows share the same appearance/identity conditioning.
\emph{The 3D representation plays a key role in maintaining smooth temporal continuity.}
While long video generation is not our main focus, using advanced techniques like temporal compression~\citep{bar2024lumiere}, beyond simple overlap, could further improve visual coherence. We leave their integration with our 3D-aware framework as future work.

\subsection{Complex Motion Generation}
\label{subsec:exp:CA}
\noindent\textbf{Experimental Setup.}
To demonstrate our model's ability to generate videos with complex motion, we compare our approach with state-of-the-art human image animation models on the TikTok Dancing dataset~\citep{jafarian2021learning}, using the Disco~\citep{wang2024disco} split.
For compatibility with the SVD model architecture, all videos are cropped to $576\times 1024$.
We fine-tune the original SVD only on the training split for 30K iterations, with a batch size of 8 and a learning rate of $1\times 10^{-5}$.

\begin{figure}
    \centering
    \includegraphics[width=1\linewidth]{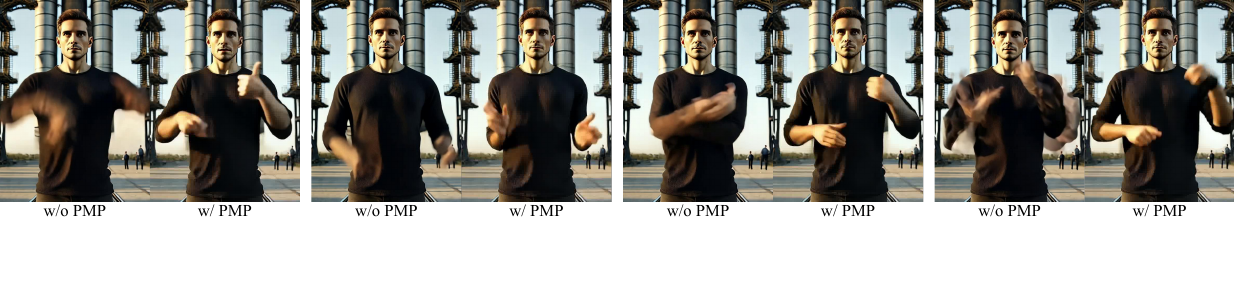}
    \vspace{-17mm}
    \caption{\motionmodelname improves motion and visual quality, generating realistic videos with large motions.
    }
    \label{fig:abl_stage1-3}
\end{figure}

\noindent\textbf{Evaluation Metrics.}
We follow baselines and report Peak Signal-to-Noise Ratio (PSNR)~\citep{hore2010image}, Structural Similarity Index (SSIM)~\citep{wang2004image}, and Learned Perceptual Image Patch Similarity (LPIPS)~\citep{zhang2018unreasonable} to measure the visual quality of the generated results.
We also report and Fréchet Video Distance (FVD)~\citep{unterthiner2018towards} for video fidelity comparision.

\noindent\textbf{Experimental Results.}
We compare \modelname with human image animation methods in Tab.~\ref{tab:dance}, where we achieve state-of-the-art performance across all metrics.
Notably, we observe a significant improvement in FVD, highlighting substantial gains in video generation quality.
It is important to note that all baselines in this task \emph{rely on ground-truth motion sequences}, which are challenging to obtain in practical scenarios, limiting their applicability. 
In contrast, our method can generate realistic and high-quality videos \emph{using only the input inference image or inference image with a target pose}.

\begin{table}[]
  \centering
  \begin{minipage}{0.44\columnwidth}
    \centering
    \caption{{\small \textbf{ Quantitative comparisons for dance generation.} `\modelname (w/ full motion)' follows baselines and takes full motion sequences as condition, while `\modelname (w/ target pose)' uses the target pose from the final frame.}}
    \vspace{-2mm}
    \setlength{\tabcolsep}{2pt}
    \resizebox{\linewidth}{!}{%
      \begin{tabular}{l|cccc}
       & {\bf SSIM $\uparrow$} & {\bf PSNR $\uparrow$} & {\bf LPIPS $\downarrow$} & {\bf FVD $\downarrow$} \\ 
      \midrule
      MagicAnimate~\citep{xu2024magicanimate} & 0.714 & 29.16 & 0.239 & 179.07 \\
      Animate Anyone~\citep{hu2024animate}   & 0.718 & 29.56 & 0.285 & 171.90 \\
      Champ~\citep{zhu2024champ}           & 0.802 & 29.91 & 0.234 & 160.82 \\
      VividPose~\citep{wang2024vividpose}     & 0.758 & 29.83 & 0.261 & 152.97 \\
      \midrule
      \modelname (image only) & - & - & - & 136.43 \\
      \modelname (w/ target pose) & - & - & - & 130.14 \\
      \modelname (w/ full motion)  & \textbf{0.864} & \textbf{30.08} & \textbf{0.210} & \textbf{121.26} \\
      \end{tabular}%
    }
    \label{tab:dance}
  \end{minipage}
  \hfill
  \begin{minipage}{0.54\columnwidth}
    \vspace{-2.5mm}
    \centering
    \caption{{\small \textbf{\motionmodelname acts as a general motion denoiser, improving performance on human motion generation.} Following MoMask, we report R-Precision at Top-1, Top-2, and Top-3. Our \motionmodelname achieves state-of-the-art performance on two widely used benchmarks.}}
    \vspace{-2mm}
    \setlength{\tabcolsep}{2pt}
    \resizebox{\linewidth}{!}{%
      \begin{tabular}{l|cccccc}
       & \multicolumn{3}{c}{{\bf HumanML3D}} & \multicolumn{3}{c}{{\bf KIT-ML}} \\
      \cmidrule(lr){2-4}\cmidrule(lr){5-7}
       & {\bf R-P@1} & {\bf R-P@2} & {\bf R-P@3} & {\bf R-P@1} & {\bf R-P@2} & {\bf R-P@3} \\
      \midrule
      MotionDiffuse~\citep{zhang2024motiondiffuse}    & 0.491 & 0.681 & 0.782 & 0.417 & 0.621 & 0.739 \\
      ReMoDiffuse~\citep{zhang2023remodiffuse}      & 0.510 & 0.698 & 0.795 & 0.427 & 0.641 & 0.765 \\
      MoMask~\citep{guo2024momask}           & 0.521 & 0.713 & 0.807 & 0.433 & 0.656 & 0.781 \\
      \midrule
      MoMask + \motionmodelname    & \textbf{0.544} & \textbf{0.735} & \textbf{0.810} 
                       & \textbf{0.471} & \textbf{0.673} & \textbf{0.785} \\
      \end{tabular}%
    }
    \label{tab:momask}
  \end{minipage}
\end{table}

\subsection{Parameterized Motion Prior Model (\motionmodelname)}
\label{subsec:exp:PPPM}
We demonstrate the effectiveness of the proposed Parameterized Motion Prior model in this section. 

\begin{wraptable}{r}{0.4\columnwidth}
  \centering
  \caption{{\small \textbf{User studies for \motionmodelname.} \motionmodelname improves object and motion consistency, while reducing morphological failure rates.}}
  \label{supp:tab:user}
  \resizebox{\linewidth}{!}{%
    \begin{tabular}{l|cc|c}
     & {\bf Object} & {\bf Motion} & {\bf  Morphological} \\
     & {\bf Consistency $\uparrow$} & {\bf Consistency $\uparrow$} & {\bf Failure Rate $\downarrow$} \\ \midrule
    w/o \motionmodelname  & 12.4 &  4.0 & 83.5 \\
    w/ \motionmodelname  & 87.6 &  96.0 & 14.3 \\
    \end{tabular}
  }
\end{wraptable}

\noindent\textbf{\motionmodelname Enables High-Quality Video Generation with Complex Motions and Interactions.}
To demonstrate the effectiveness of \motionmodelname, we select a complex dance scenario and visualize outputs with and without the proposed \motionmodelname in Fig.~\ref{fig:abl_stage1-3}.
We also show quantitative improvements of the generated videos with \motionmodelname in Tab.~\ref{supp:tab:user}, where 500 video pairs were evaluated by random users on Amazon MTurk. Each pair was rated by three different users, resulting in a total of 1,500 evaluations.
The results show that the video generation model alone still struggles to produce high-quality videos with accurate motion.
However, \emph{leveraging the object-level priors from our Parameterized Motion Prior enables the generation of realistic videos with enhanced motion and visual quality.}

\noindent\textbf{\motionmodelname Prevents Error Accumulation in Multi-stage Video Generation.} In addition, Fig.~\ref{fig:abl_stage1-3} shows that even when the generated videos exhibit severely broken motion, our \motionmodelname can still recover (predict) a smooth and coherent motion sequence using the ground-truth first frame and target pose, enabling successful final video generation. \emph{This correction mitigates motion errors and prevents error accumulation, highlighting the robustness of our pipeline}.

\noindent\textbf{\motionmodelname as a General Motion Denoiser.} 
We focus on a more specific human motion generation task and show that our model improves the performance of the state-of-the-art method, MoMask~\citep{guo2024momask}, on standard benchmarks (see Tab.~\ref{tab:momask}). 
Specifically, we use \motionmodelname to refine the motion sequences generated by MoMask and compare the results with MoMask and other methods on HumanML3D~\citep{guo2022generating} and KIT-ML~\citep{plappert2016kit} benchmarks. 
We adopt the same training dataset as MoMask and apply the perturbations described in Sec.~\ref{subsec:revision} to train \motionmodelname. 
\emph{Serving as a general motion denoiser, our model consistently enhances motion generation quality of the current best models across multiple benchmarks.}

\section{Conclusion}
We introduced \modelname, a three-stage framework for video generation that improves motion consistency by integrating 3D motion cues. 
\modelname leverages a pretrained video diffusion model to generate coarse videos, refines 3D motion sequences with \motionmodelname, and reconditions the generation process with enhanced motions to improve fine-grained and complex motion generations.
Evaluations show that \modelname significantly outperforms existing methods in motion fidelity and coherence.

\noindent\textbf{Acknowledgements.}
Thanks to all who supported this project and to the anonymous reviewers for their constructive comments.
AY acknowledges support from the ONR N000142412696.

\newpage

\bibliography{main}

\begin{thebibliography}{95}
\providecommand{\natexlab}[1]{#1}
\providecommand{\url}[1]{\texttt{#1}}
\expandafter\ifx\csname urlstyle\endcsname\relax
  \providecommand{\doi}[1]{doi: #1}\else
  \providecommand{\doi}{doi: \begingroup \urlstyle{rm}\Url}\fi

\bibitem[Babaeizadeh et~al.(2017)Babaeizadeh, Finn, Erhan, Campbell, and Levine]{babaeizadeh2017stochastic}
Mohammad Babaeizadeh, Chelsea Finn, Dumitru Erhan, Roy~H Campbell, and Sergey Levine.
\newblock Stochastic variational video prediction.
\newblock \emph{arXiv preprint arXiv:1710.11252}, 2017.

\bibitem[Bain et~al.(2021)Bain, Nagrani, Varol, and Zisserman]{bain2021frozen}
Max Bain, Arsha Nagrani, G{\"u}l Varol, and Andrew Zisserman.
\newblock Frozen in time: A joint video and image encoder for end-to-end retrieval.
\newblock In \emph{Proceedings of the IEEE/CVF International Conference on Computer Vision}, pp.\  1728--1738, 2021.

\bibitem[Bao et~al.(2023)Bao, Nie, Xue, Cao, Li, Su, and Zhu]{bao2023all}
Fan Bao, Shen Nie, Kaiwen Xue, Yue Cao, Chongxuan Li, Hang Su, and Jun Zhu.
\newblock All are worth words: A vit backbone for diffusion models.
\newblock In \emph{Proceedings of the IEEE/CVF Conference on Computer Vision and Pattern Recognition}, pp.\  22669--22679, 2023.

\bibitem[Bar-Tal et~al.(2024)Bar-Tal, Chefer, Tov, Herrmann, Paiss, Zada, Ephrat, Hur, Liu, Raj, et~al.]{bar2024lumiere}
Omer Bar-Tal, Hila Chefer, Omer Tov, Charles Herrmann, Roni Paiss, Shiran Zada, Ariel Ephrat, Junhwa Hur, Guanghui Liu, Amit Raj, et~al.
\newblock Lumiere: A space-time diffusion model for video generation.
\newblock In \emph{SIGGRAPH Asia 2024 Conference Papers}, pp.\  1--11, 2024.

\bibitem[Betker et~al.(2023)Betker, Goh, Jing, Brooks, Wang, Li, Ouyang, Zhuang, Lee, Guo, et~al.]{betker2023improving}
James Betker, Gabriel Goh, Li~Jing, Tim Brooks, Jianfeng Wang, Linjie Li, Long Ouyang, Juntang Zhuang, Joyce Lee, Yufei Guo, et~al.
\newblock Improving image generation with better captions.
\newblock \emph{Computer Science. https://cdn. openai. com/papers/dall-e-3. pdf}, 2\penalty0 (3):\penalty0 8, 2023.

\bibitem[Blattmann et~al.(2023)Blattmann, Dockhorn, Kulal, Mendelevitch, Kilian, Lorenz, Levi, English, Voleti, Letts, et~al.]{blattmann2023stable}
Andreas Blattmann, Tim Dockhorn, Sumith Kulal, Daniel Mendelevitch, Maciej Kilian, Dominik Lorenz, Yam Levi, Zion English, Vikram Voleti, Adam Letts, et~al.
\newblock Stable video diffusion: Scaling latent video diffusion models to large datasets.
\newblock \emph{arXiv preprint arXiv:2311.15127}, 2023.

\bibitem[Brooks et~al.(2024)Brooks, Peebles, Holmes, DePue, Guo, Jing, Schnurr, Taylor, Luhman, Luhman, et~al.]{brooks2024video}
Tim Brooks, Bill Peebles, Connor Holmes, Will DePue, Yufei Guo, Li~Jing, David Schnurr, Joe Taylor, Troy Luhman, Eric Luhman, et~al.
\newblock Video generation models as world simulators, 2024.

\bibitem[Burgert et~al.(2025)Burgert, Xu, Xian, Pilarski, Clausen, He, Ma, Deng, Li, Mousavi, et~al.]{burgert2025go}
Ryan Burgert, Yuancheng Xu, Wenqi Xian, Oliver Pilarski, Pascal Clausen, Mingming He, Li~Ma, Yitong Deng, Lingxiao Li, Mohsen Mousavi, et~al.
\newblock Go-with-the-flow: Motion-controllable video diffusion models using real-time warped noise.
\newblock In \emph{Proceedings of the Computer Vision and Pattern Recognition Conference}, pp.\  13--23, 2025.

\bibitem[Chen et~al.(2024)Chen, Siarohin, Menapace, Deyneka, Chao, Jeon, Fang, Lee, Ren, Yang, et~al.]{chen2024panda}
Tsai-Shien Chen, Aliaksandr Siarohin, Willi Menapace, Ekaterina Deyneka, Hsiang-wei Chao, Byung~Eun Jeon, Yuwei Fang, Hsin-Ying Lee, Jian Ren, Ming-Hsuan Yang, et~al.
\newblock Panda-70m: Captioning 70m videos with multiple cross-modality teachers.
\newblock In \emph{Proceedings of the IEEE/CVF Conference on Computer Vision and Pattern Recognition}, pp.\  13320--13331, 2024.

\bibitem[Chen et~al.(2023)Chen, Wang, Zhang, Zhuang, Ma, Yu, Wang, Lin, Qiao, and Liu]{chen2023seine}
Xinyuan Chen, Yaohui Wang, Lingjun Zhang, Shaobin Zhuang, Xin Ma, Jiashuo Yu, Yali Wang, Dahua Lin, Yu~Qiao, and Ziwei Liu.
\newblock Seine: Short-to-long video diffusion model for generative transition and prediction.
\newblock In \emph{The Twelfth International Conference on Learning Representations}, 2023.

\bibitem[Esser et~al.(2024)Esser, Kulal, Blattmann, Entezari, M{\"u}ller, Saini, Levi, Lorenz, Sauer, Boesel, et~al.]{esser2024scaling}
Patrick Esser, Sumith Kulal, Andreas Blattmann, Rahim Entezari, Jonas M{\"u}ller, Harry Saini, Yam Levi, Dominik Lorenz, Axel Sauer, Frederic Boesel, et~al.
\newblock Scaling rectified flow transformers for high-resolution image synthesis.
\newblock In \emph{Forty-first International Conference on Machine Learning}, 2024.

\bibitem[Girdhar et~al.(2023)Girdhar, Singh, Brown, Duval, Azadi, Rambhatla, Shah, Yin, Parikh, and Misra]{girdhar2023emu}
Rohit Girdhar, Mannat Singh, Andrew Brown, Quentin Duval, Samaneh Azadi, Sai~Saketh Rambhatla, Akbar Shah, Xi~Yin, Devi Parikh, and Ishan Misra.
\newblock Emu video: Factorizing text-to-video generation by explicit image conditioning.
\newblock \emph{arXiv preprint arXiv:2311.10709}, 2023.

\bibitem[Goel et~al.(2023)Goel, Pavlakos, Rajasegaran, Kanazawa, and Malik]{goel2023humans}
Shubham Goel, Georgios Pavlakos, Jathushan Rajasegaran, Angjoo Kanazawa, and Jitendra Malik.
\newblock Humans in 4d: Reconstructing and tracking humans with transformers.
\newblock In \emph{Proceedings of the IEEE/CVF International Conference on Computer Vision}, pp.\  14783--14794, 2023.

\bibitem[Guo et~al.(2022)Guo, Zou, Zuo, Wang, Ji, Li, and Cheng]{guo2022generating}
Chuan Guo, Shihao Zou, Xinxin Zuo, Sen Wang, Wei Ji, Xingyu Li, and Li~Cheng.
\newblock Generating diverse and natural 3d human motions from text.
\newblock In \emph{Proceedings of the IEEE/CVF conference on computer vision and pattern recognition}, pp.\  5152--5161, 2022.

\bibitem[Guo et~al.(2024)Guo, Mu, Javed, Wang, and Cheng]{guo2024momask}
Chuan Guo, Yuxuan Mu, Muhammad~Gohar Javed, Sen Wang, and Li~Cheng.
\newblock Momask: Generative masked modeling of 3d human motions.
\newblock In \emph{Proceedings of the IEEE/CVF Conference on Computer Vision and Pattern Recognition}, pp.\  1900--1910, 2024.

\bibitem[Ho et~al.(2020)Ho, Jain, and Abbeel]{ho2020denoising}
Jonathan Ho, Ajay Jain, and Pieter Abbeel.
\newblock Denoising diffusion probabilistic models.
\newblock \emph{Advances in Neural Information Processing Systems}, 33:\penalty0 6840--6851, 2020.

\bibitem[Ho et~al.(2022)Ho, Salimans, Gritsenko, Chan, Norouzi, and Fleet]{ho2022video}
Jonathan Ho, Tim Salimans, Alexey Gritsenko, William Chan, Mohammad Norouzi, and David~J Fleet.
\newblock Video diffusion models.
\newblock \emph{Advances in Neural Information Processing Systems}, 35:\penalty0 8633--8646, 2022.

\bibitem[Hore \& Ziou(2010)Hore and Ziou]{hore2010image}
Alain Hore and Djemel Ziou.
\newblock Image quality metrics: Psnr vs. ssim.
\newblock In \emph{2010 20th International Conference on Pattern Recognition}, pp.\  2366--2369, 2010.

\bibitem[Hu(2024)]{hu2024animate}
Li~Hu.
\newblock Animate anyone: Consistent and controllable image-to-video synthesis for character animation.
\newblock In \emph{Proceedings of the IEEE/CVF Conference on Computer Vision and Pattern Recognition}, pp.\  8153--8163, 2024.

\bibitem[Hu et~al.(2022)Hu, Luo, and Chen]{hu2022make}
Yaosi Hu, Chong Luo, and Zhenzhong Chen.
\newblock Make it move: controllable image-to-video generation with text descriptions.
\newblock In \emph{Proceedings of the IEEE/CVF Conference on Computer Vision and Pattern Recognition}, pp.\  18219--18228, 2022.

\bibitem[Huang et~al.(2025)Huang, Ma, Duan, Chen, Wan, Ming, Wang, Wang, Lu, Li, et~al.]{huang2025step}
Haoyang Huang, Guoqing Ma, Nan Duan, Xing Chen, Changyi Wan, Ranchen Ming, Tianyu Wang, Bo~Wang, Zhiying Lu, Aojie Li, et~al.
\newblock Step-video-ti2v technical report: A state-of-the-art text-driven image-to-video generation model.
\newblock \emph{arXiv preprint arXiv:2503.11251}, 2025.

\bibitem[Huang et~al.(2024)Huang, Zhang, Xu, He, Yu, Dong, Ma, Chanpaisit, Si, Jiang, et~al.]{huang2024vbench}
Ziqi Huang, Fan Zhang, Xiaojie Xu, Yinan He, Jiashuo Yu, Ziyue Dong, Qianli Ma, Nattapol Chanpaisit, Chenyang Si, Yuming Jiang, et~al.
\newblock Vbench++: Comprehensive and versatile benchmark suite for video generative models.
\newblock \emph{arXiv preprint arXiv:2411.13503}, 2024.

\bibitem[Hurst et~al.(2024)Hurst, Lerer, Goucher, Perelman, Ramesh, Clark, Ostrow, Welihinda, Hayes, Radford, et~al.]{hurst2024gpt}
Aaron Hurst, Adam Lerer, Adam~P Goucher, Adam Perelman, Aditya Ramesh, Aidan Clark, AJ~Ostrow, Akila Welihinda, Alan Hayes, Alec Radford, et~al.
\newblock Gpt-4o system card.
\newblock \emph{arXiv preprint arXiv:2410.21276}, 2024.

\bibitem[Jafarian \& Park(2021)Jafarian and Park]{jafarian2021learning}
Yasamin Jafarian and Hyun~Soo Park.
\newblock Learning high fidelity depths of dressed humans by watching social media dance videos.
\newblock In \emph{Proceedings of the IEEE/CVF Conference on Computer Vision and Pattern Recognition}, pp.\  12753--12762, 2021.

\bibitem[Jiang et~al.(2023)Jiang, Lu, Zhang, Ma, Han, Lyu, Li, and Chen]{jiang2023rtmpose}
Tao Jiang, Peng Lu, Li~Zhang, Ningsheng Ma, Rui Han, Chengqi Lyu, Yining Li, and Kai Chen.
\newblock Rtmpose: Real-time multi-person pose estimation based on mmpose.
\newblock \emph{arXiv preprint arXiv:2303.07399}, 2023.

\bibitem[Kang et~al.(2024)Kang, Yue, Lu, Lin, Zhao, Wang, Huang, and Feng]{kang2024far}
Bingyi Kang, Yang Yue, Rui Lu, Zhijie Lin, Yang Zhao, Kaixin Wang, Gao Huang, and Jiashi Feng.
\newblock How far is video generation from world model: A physical law perspective.
\newblock \emph{arXiv preprint arXiv:2411.02385}, 2024.

\bibitem[Kingma \& Welling(2014)Kingma and Welling]{kingma2013auto}
Diederik~P Kingma and Max Welling.
\newblock Auto-encoding variational bayes.
\newblock In \emph{International Conference on Learning Representations}, 2014.

\bibitem[Kirillov et~al.(2023)Kirillov, Mintun, Ravi, Mao, Rolland, Gustafson, Xiao, Whitehead, Berg, Lo, Doll{\'a}r, and Girshick]{kirillov2023segany}
Alexander Kirillov, Eric Mintun, Nikhila Ravi, Hanzi Mao, Chloe Rolland, Laura Gustafson, Tete Xiao, Spencer Whitehead, Alexander~C. Berg, Wan-Yen Lo, Piotr Doll{\'a}r, and Ross Girshick.
\newblock Segment anything.
\newblock \emph{arXiv:2304.02643}, 2023.

\bibitem[Kong et~al.(2024)Kong, Tian, Zhang, Min, Dai, Zhou, Xiong, Li, Wu, Zhang, et~al.]{kong2024hunyuanvideo}
Weijie Kong, Qi~Tian, Zijian Zhang, Rox Min, Zuozhuo Dai, Jin Zhou, Jiangfeng Xiong, Xin Li, Bo~Wu, Jianwei Zhang, et~al.
\newblock Hunyuanvideo: A systematic framework for large video generative models.
\newblock \emph{arXiv preprint arXiv:2412.03603}, 2024.

\bibitem[Li et~al.(2025)Li, Wang, Zhang, Wang, Yuan, Xie, Shan, and Zou]{li2025image}
Yaowei Li, Xintao Wang, Zhaoyang Zhang, Zhouxia Wang, Ziyang Yuan, Liangbin Xie, Ying Shan, and Yuexian Zou.
\newblock Image conductor: Precision control for interactive video synthesis.
\newblock In \emph{Proceedings of the AAAI Conference on Artificial Intelligence}, volume~39, pp.\  5031--5038, 2025.

\bibitem[Li et~al.(2018)Li, Fang, Yang, Wang, Lu, and Yang]{li2018flow}
Yijun Li, Chen Fang, Jimei Yang, Zhaowen Wang, Xin Lu, and Ming-Hsuan Yang.
\newblock Flow-grounded spatial-temporal video prediction from still images.
\newblock In \emph{Proceedings of the European Conference on Computer Vision (ECCV)}, pp.\  600--615, 2018.

\bibitem[Ling et~al.(2024)Ling, Bu, Zhang, Dong, Zang, Wu, Chen, Wang, and Jin]{ling2024motionclone}
Pengyang Ling, Jiazi Bu, Pan Zhang, Xiaoyi Dong, Yuhang Zang, Tong Wu, Huaian Chen, Jiaqi Wang, and Yi~Jin.
\newblock Motionclone: Training-free motion cloning for controllable video generation.
\newblock \emph{arXiv preprint arXiv:2406.05338}, 2024.

\bibitem[Lipman et~al.(2022)Lipman, Chen, Ben-Hamu, Nickel, and Le]{lipman2022flow}
Yaron Lipman, Ricky~TQ Chen, Heli Ben-Hamu, Maximilian Nickel, and Matt Le.
\newblock Flow matching for generative modeling.
\newblock \emph{arXiv preprint arXiv:2210.02747}, 2022.

\bibitem[Liu et~al.(2024{\natexlab{a}})Liu, Akhgari, Visheratin, Kamko, Xu, Shrirao, Lambert, Souza, Doshi, and Li]{liu2024playground}
Bingchen Liu, Ehsan Akhgari, Alexander Visheratin, Aleks Kamko, Linmiao Xu, Shivam Shrirao, Chase Lambert, Joao Souza, Suhail Doshi, and Daiqing Li.
\newblock Playground v3: Improving text-to-image alignment with deep-fusion large language models.
\newblock \emph{arXiv preprint arXiv:2409.10695}, 2024{\natexlab{a}}.

\bibitem[Liu et~al.(2024{\natexlab{b}})Liu, Yin, Yuille, Brown, and Singh]{liu2024flowing}
Qihao Liu, Xi~Yin, Alan Yuille, Andrew Brown, and Mannat Singh.
\newblock Flowing from words to pixels: A framework for cross-modality evolution.
\newblock \emph{arXiv preprint arXiv:2412.15213}, 2024{\natexlab{b}}.

\bibitem[Liu et~al.(2024{\natexlab{c}})Liu, Zeng, He, Yu, Shen, and Chen]{liu2024alleviating}
Qihao Liu, Zhanpeng Zeng, Ju~He, Qihang Yu, Xiaohui Shen, and Liang-Chieh Chen.
\newblock Alleviating distortion in image generation via multi-resolution diffusion models and time-dependent layer normalization.
\newblock \emph{Advances in Neural Information Processing Systems}, 37:\penalty0 133879--133907, 2024{\natexlab{c}}.

\bibitem[Liu et~al.(2024{\natexlab{d}})Liu, Zhang, Bai, Kortylewski, and Yuille]{liu2024direct}
Qihao Liu, Yi~Zhang, Song Bai, Adam Kortylewski, and Alan Yuille.
\newblock Direct-3d: Learning direct text-to-3d generation on massive noisy 3d data.
\newblock In \emph{Proceedings of the IEEE/CVF Conference on Computer Vision and Pattern Recognition}, pp.\  6881--6891, 2024{\natexlab{d}}.

\bibitem[Liu et~al.(2022)Liu, Gong, and Liu]{liu2022flow}
Xingchao Liu, Chengyue Gong, and Qiang Liu.
\newblock Flow straight and fast: Learning to generate and transfer data with rectified flow.
\newblock \emph{arXiv preprint arXiv:2209.03003}, 2022.

\bibitem[Loper et~al.(2015)Loper, Mahmood, Romero, Pons-Moll, and Black]{loper2023smpl}
Matthew Loper, Naureen Mahmood, Javier Romero, Gerard Pons-Moll, and Michael~J. Black.
\newblock {SMPL}: A skinned multi-person linear model.
\newblock \emph{ACM Trans. Graphics (Proc. SIGGRAPH Asia)}, 34\penalty0 (6):\penalty0 248:1--248:16, October 2015.

\bibitem[Ma et~al.(2024)Ma, Goldstein, Albergo, Boffi, Vanden-Eijnden, and Xie]{ma2024sit}
Nanye Ma, Mark Goldstein, Michael~S Albergo, Nicholas~M Boffi, Eric Vanden-Eijnden, and Saining Xie.
\newblock Sit: Exploring flow and diffusion-based generative models with scalable interpolant transformers.
\newblock In \emph{European Conference on Computer Vision}, pp.\  23--40. Springer, 2024.

\bibitem[Mahapatra \& Kulkarni(2022)Mahapatra and Kulkarni]{mahapatra2022controllable}
Aniruddha Mahapatra and Kuldeep Kulkarni.
\newblock Controllable animation of fluid elements in still images.
\newblock In \emph{Proceedings of the IEEE/CVF Conference on Computer Vision and Pattern Recognition}, pp.\  3667--3676, 2022.

\bibitem[Ni et~al.(2023)Ni, Shi, Li, Huang, and Min]{ni2023conditional}
Haomiao Ni, Changhao Shi, Kai Li, Sharon~X Huang, and Martin~Renqiang Min.
\newblock Conditional image-to-video generation with latent flow diffusion models.
\newblock In \emph{Proceedings of the IEEE/CVF Conference on Computer Vision and Pattern Recognition}, pp.\  18444--18455, 2023.

\bibitem[Pan et~al.(2019)Pan, Wang, Jia, Shao, Sheng, Yan, and Wang]{pan2019video}
Junting Pan, Chengyu Wang, Xu~Jia, Jing Shao, Lu~Sheng, Junjie Yan, and Xiaogang Wang.
\newblock Video generation from single semantic label map.
\newblock In \emph{Proceedings of the IEEE/CVF Conference on Computer Vision and Pattern Recognition}, pp.\  3733--3742, 2019.

\bibitem[Pavlakos et~al.(2019)Pavlakos, Choutas, Ghorbani, Bolkart, Osman, Tzionas, and Black]{pavlakos2019expressive}
Georgios Pavlakos, Vasileios Choutas, Nima Ghorbani, Timo Bolkart, Ahmed~AA Osman, Dimitrios Tzionas, and Michael~J Black.
\newblock Expressive body capture: 3d hands, face, and body from a single image.
\newblock In \emph{Proceedings of the IEEE/CVF Conference on Computer Vision and Pattern Recognition}, pp.\  10975--10985, 2019.

\bibitem[Pavlakos et~al.(2024)Pavlakos, Shan, Radosavovic, Kanazawa, Fouhey, and Malik]{pavlakos2024reconstructing}
Georgios Pavlakos, Dandan Shan, Ilija Radosavovic, Angjoo Kanazawa, David Fouhey, and Jitendra Malik.
\newblock Reconstructing hands in 3d with transformers.
\newblock In \emph{Proceedings of the IEEE/CVF Conference on Computer Vision and Pattern Recognition}, pp.\  9826--9836, 2024.

\bibitem[Peebles \& Xie(2023)Peebles and Xie]{peebles2023scalable}
William Peebles and Saining Xie.
\newblock Scalable diffusion models with transformers.
\newblock In \emph{Proceedings of the IEEE/CVF International Conference on Computer Vision}, pp.\  4195--4205, 2023.

\bibitem[Peng et~al.(2020)Peng, Jiang, Pi, Li, Bao, and Zhou]{peng2020deep}
Sida Peng, Wen Jiang, Huaijin Pi, Xiuli Li, Hujun Bao, and Xiaowei Zhou.
\newblock Deep snake for real-time instance segmentation.
\newblock In \emph{Proceedings of the IEEE/CVF Conference on Computer Vision and Pattern Recognition}, pp.\  8533--8542, 2020.

\bibitem[Plappert et~al.(2016)Plappert, Mandery, and Asfour]{plappert2016kit}
Matthias Plappert, Christian Mandery, and Tamim Asfour.
\newblock The kit motion-language dataset.
\newblock \emph{Big data}, 4\penalty0 (4):\penalty0 236--252, 2016.

\bibitem[Polyak et~al.(2024)Polyak, Zohar, Brown, Tjandra, Sinha, Lee, Vyas, Shi, et~al.]{polyak2024moviegencastmedia}
Adam Polyak, Amit Zohar, Andrew Brown, Andros Tjandra, Animesh Sinha, Ann Lee, Apoorv Vyas, Bowen Shi, et~al.
\newblock Movie gen: A cast of media foundation models, 2024.
\newblock URL \url{https://arxiv.org/abs/2410.13720}.

\bibitem[Pont-Tuset et~al.(2017)Pont-Tuset, Perazzi, Caelles, Arbel{\'a}ez, Sorkine-Hornung, and Van~Gool]{pont20172017}
Jordi Pont-Tuset, Federico Perazzi, Sergi Caelles, Pablo Arbel{\'a}ez, Alex Sorkine-Hornung, and Luc Van~Gool.
\newblock The 2017 davis challenge on video object segmentation.
\newblock \emph{arXiv preprint arXiv:1704.00675}, 2017.

\bibitem[Radford et~al.(2021)Radford, Kim, Hallacy, Ramesh, Goh, Agarwal, Sastry, Askell, Mishkin, Clark, et~al.]{radford2021learning}
Alec Radford, Jong~Wook Kim, Chris Hallacy, Aditya Ramesh, Gabriel Goh, Sandhini Agarwal, Girish Sastry, Amanda Askell, Pamela Mishkin, Jack Clark, et~al.
\newblock Learning transferable visual models from natural language supervision.
\newblock In \emph{International Conference on Machine Learning}, 2021.

\bibitem[Ravi et~al.(2024)Ravi, Gabeur, Hu, Hu, Ryali, Ma, Khedr, Rädle, Rolland, Gustafson, Mintun, Pan, Alwala, Carion, Wu, Girshick, Dollár, and Feichtenhofer]{ravi2024sam2segmentimages}
Nikhila Ravi, Valentin Gabeur, Yuan-Ting Hu, Ronghang Hu, Chaitanya Ryali, Tengyu Ma, Haitham Khedr, Roman Rädle, Chloe Rolland, Laura Gustafson, Eric Mintun, Junting Pan, Kalyan~Vasudev Alwala, Nicolas Carion, Chao-Yuan Wu, Ross Girshick, Piotr Dollár, and Christoph Feichtenhofer.
\newblock Sam 2: Segment anything in images and videos, 2024.
\newblock URL \url{https://arxiv.org/abs/2408.00714}.

\bibitem[Ren et~al.(2024{\natexlab{a}})Ren, Liu, Zeng, Lin, Li, Cao, Chen, Huang, Chen, Yan, Zeng, Zhang, Li, Yang, Li, Jiang, and Zhang]{ren2024grounded}
Tianhe Ren, Shilong Liu, Ailing Zeng, Jing Lin, Kunchang Li, He~Cao, Jiayu Chen, Xinyu Huang, Yukang Chen, Feng Yan, Zhaoyang Zeng, Hao Zhang, Feng Li, Jie Yang, Hongyang Li, Qing Jiang, and Lei Zhang.
\newblock Grounded sam: Assembling open-world models for diverse visual tasks, 2024{\natexlab{a}}.

\bibitem[Ren et~al.(2024{\natexlab{b}})Ren, Yang, Zhang, Wei, Du, Huang, and Chen]{ren2024consisti2v}
Weiming Ren, Huan Yang, Ge~Zhang, Cong Wei, Xinrun Du, Wenhao Huang, and Wenhu Chen.
\newblock Consisti2v: Enhancing visual consistency for image-to-video generation.
\newblock \emph{arXiv preprint arXiv:2402.04324}, 2024{\natexlab{b}}.

\bibitem[Rombach et~al.(2022)Rombach, Blattmann, Lorenz, Esser, and Ommer]{rombach2022high}
Robin Rombach, Andreas Blattmann, Dominik Lorenz, Patrick Esser, and Bj{\"o}rn Ommer.
\newblock High-resolution image synthesis with latent diffusion models.
\newblock In \emph{Proceedings IEEE/CVF Conference on Computer Vision and Pattern Recognition (CVPR)}, 2022.

\bibitem[Romero et~al.(2022)Romero, Tzionas, and Black]{romero2022embodied}
Javier Romero, Dimitrios Tzionas, and Michael~J Black.
\newblock Embodied hands: Modeling and capturing hands and bodies together.
\newblock \emph{arXiv preprint arXiv:2201.02610}, 2022.

\bibitem[Ronneberger et~al.(2015)Ronneberger, Fischer, and Brox]{ronneberger2015u}
Olaf Ronneberger, Philipp Fischer, and Thomas Brox.
\newblock U-net: Convolutional networks for biomedical image segmentation.
\newblock In \emph{Medical image computing and computer-assisted intervention--MICCAI 2015: 18th international conference, Munich, Germany, October 5-9, 2015, proceedings, part III 18}, pp.\  234--241, 2015.

\bibitem[Rueegg et~al.(2023)Rueegg, Tripathi, Schindler, Black, and Zuffi]{rueegg2023bite}
Nadine Rueegg, Shashank Tripathi, Konrad Schindler, Michael~J. Black, and Silvia Zuffi.
\newblock Bite: Beyond priors for improved three-d dog pose estimation.
\newblock In \emph{Proceedings IEEE Conferecne on Computer Vision and Pattern Recognition (CVPR)}, 2023.

\bibitem[{Runway}(2026)]{gen4}
{Runway}.
\newblock Runwayml – ai video and image generator, 2026.
\newblock URL \url{https://runwayml.com/}.

\bibitem[Sabathier et~al.(2024)Sabathier, Mitra, and Novotny]{Sabathier2024AnimalAR}
Remy Sabathier, Niloy~Jyoti Mitra, and David Novotny.
\newblock Animal avatars: Reconstructing animatable {3D} animals from casual videos.
\newblock \emph{ArXiv}, abs/2403.17103, 2024.

\bibitem[Shi et~al.(2024)Shi, Huang, Wang, Bian, Li, Zhang, Zhang, Cheung, See, Qin, et~al.]{shi2024motion}
Xiaoyu Shi, Zhaoyang Huang, Fu-Yun Wang, Weikang Bian, Dasong Li, Yi~Zhang, Manyuan Zhang, Ka~Chun Cheung, Simon See, Hongwei Qin, et~al.
\newblock Motion-i2v: Consistent and controllable image-to-video generation with explicit motion modeling.
\newblock In \emph{ACM SIGGRAPH 2024 Conference Papers}, pp.\  1--11, 2024.

\bibitem[Singer et~al.(2022)Singer, Polyak, Hayes, Yin, An, Zhang, Hu, Yang, Ashual, Gafni, et~al.]{singer2022make}
Uriel Singer, Adam Polyak, Thomas Hayes, Xi~Yin, Jie An, Songyang Zhang, Qiyuan Hu, Harry Yang, Oron Ashual, Oran Gafni, et~al.
\newblock Make-a-video: Text-to-video generation without text-video data.
\newblock \emph{arXiv preprint arXiv:2209.14792}, 2022.

\bibitem[Song et~al.(2020)Song, Sohl-Dickstein, Kingma, Kumar, Ermon, and Poole]{song2020score}
Yang Song, Jascha Sohl-Dickstein, Diederik~P Kingma, Abhishek Kumar, Stefano Ermon, and Ben Poole.
\newblock Score-based generative modeling through stochastic differential equations.
\newblock \emph{arXiv preprint arXiv:2011.13456}, 2020.

\bibitem[Sun et~al.(2019)Sun, Xiao, Liu, and Wang]{sun2019deep}
Ke~Sun, Bin Xiao, Dong Liu, and Jingdong Wang.
\newblock Deep high-resolution representation learning for human pose estimation.
\newblock In \emph{Proceedings of the IEEE/CVF Conference on Computer Vision and Pattern Recognition}, pp.\  5693--5703, 2019.

\bibitem[Tan et~al.(2024)Tan, Gong, Wang, Zhang, Zheng, Zheng, Zheng, Chen, and Yang]{tan2024animate}
Shuai Tan, Biao Gong, Xiang Wang, Shiwei Zhang, Dandan Zheng, Ruobing Zheng, Kecheng Zheng, Jingdong Chen, and Ming Yang.
\newblock Animate-x: Universal character image animation with enhanced motion representation.
\newblock \emph{arXiv preprint arXiv:2410.10306}, 2024.

\bibitem[Teng et~al.(2025)Teng, Jia, Sun, Li, Li, Tang, Han, Zhang, Zhang, Luo, et~al.]{teng2025magi}
Hansi Teng, Hongyu Jia, Lei Sun, Lingzhi Li, Maolin Li, Mingqiu Tang, Shuai Han, Tianning Zhang, WQ~Zhang, Weifeng Luo, et~al.
\newblock Magi-1: Autoregressive video generation at scale.
\newblock \emph{arXiv preprint arXiv:2505.13211}, 2025.

\bibitem[Unterthiner et~al.(2018)Unterthiner, Van~Steenkiste, Kurach, Marinier, Michalski, and Gelly]{unterthiner2018towards}
Thomas Unterthiner, Sjoerd Van~Steenkiste, Karol Kurach, Raphael Marinier, Marcin Michalski, and Sylvain Gelly.
\newblock Towards accurate generative models of video: A new metric \& challenges.
\newblock \emph{arXiv preprint arXiv:1812.01717}, 2018.

\bibitem[Varghese \& Sambath(2024)Varghese and Sambath]{varghese2024yolov8}
Rejin Varghese and M~Sambath.
\newblock Yolov8: A novel object detection algorithm with enhanced performance and robustness.
\newblock In \emph{2024 International Conference on Advances in Data Engineering and Intelligent Computing Systems (ADICS)}, pp.\  1--6. IEEE, 2024.

\bibitem[Vaswani(2017)]{vaswani2017attention}
A~Vaswani.
\newblock Attention is all you need.
\newblock \emph{Advances in Neural Information Processing Systems}, 2017.

\bibitem[Wan et~al.(2025)Wan, Wang, Ai, Wen, Mao, Xie, Chen, Yu, Zhao, Yang, Zeng, Wang, Zhang, Zhou, Wang, Chen, Zhu, Zhao, Yan, Huang, Feng, Zhang, Li, Wu, Chu, Feng, Zhang, Sun, Fang, Wang, Gui, Weng, Shen, Lin, Wang, Wang, Zhou, Wang, Shen, Yu, Shi, Huang, Xu, Kou, Lv, Li, Liu, Wang, Zhang, Huang, Li, Wu, Liu, Pan, Zheng, Hong, Shi, Feng, Jiang, Han, Wu, and Liu]{wan2025}
Team Wan, Ang Wang, Baole Ai, Bin Wen, Chaojie Mao, Chen-Wei Xie, Di~Chen, Feiwu Yu, Haiming Zhao, Jianxiao Yang, Jianyuan Zeng, Jiayu Wang, Jingfeng Zhang, Jingren Zhou, Jinkai Wang, Jixuan Chen, Kai Zhu, Kang Zhao, Keyu Yan, Lianghua Huang, Mengyang Feng, Ningyi Zhang, Pandeng Li, Pingyu Wu, Ruihang Chu, Ruili Feng, Shiwei Zhang, Siyang Sun, Tao Fang, Tianxing Wang, Tianyi Gui, Tingyu Weng, Tong Shen, Wei Lin, Wei Wang, Wei Wang, Wenmeng Zhou, Wente Wang, Wenting Shen, Wenyuan Yu, Xianzhong Shi, Xiaoming Huang, Xin Xu, Yan Kou, Yangyu Lv, Yifei Li, Yijing Liu, Yiming Wang, Yingya Zhang, Yitong Huang, Yong Li, You Wu, Yu~Liu, Yulin Pan, Yun Zheng, Yuntao Hong, Yupeng Shi, Yutong Feng, Zeyinzi Jiang, Zhen Han, Zhi-Fan Wu, and Ziyu Liu.
\newblock Wan: Open and advanced large-scale video generative models.
\newblock \emph{arXiv preprint arXiv:2503.20314}, 2025.

\bibitem[Wang et~al.(2024{\natexlab{a}})Wang, Karaev, Rupprecht, and Novotny]{wang2024vggsfm}
Jianyuan Wang, Nikita Karaev, Christian Rupprecht, and David Novotny.
\newblock Vggsfm: Visual geometry grounded deep structure from motion.
\newblock In \emph{Proceedings of the IEEE/CVF Conference on Computer Vision and Pattern Recognition}, pp.\  21686--21697, 2024{\natexlab{a}}.

\bibitem[Wang et~al.(2024{\natexlab{b}})Wang, Jiang, Xu, Zhang, Wang, Zhang, Cao, Cao, Wang, and Fu]{wang2024vividpose}
Qilin Wang, Zhengkai Jiang, Chengming Xu, Jiangning Zhang, Yabiao Wang, Xinyi Zhang, Yun Cao, Weijian Cao, Chengjie Wang, and Yanwei Fu.
\newblock Vividpose: Advancing stable video diffusion for realistic human image animation.
\newblock \emph{arXiv preprint arXiv:2405.18156}, 2024{\natexlab{b}}.

\bibitem[Wang et~al.(2024{\natexlab{c}})Wang, Li, Lin, Zhai, Lin, Yang, Zhang, Liu, and Wang]{wang2024disco}
Tan Wang, Linjie Li, Kevin Lin, Yuanhao Zhai, Chung-Ching Lin, Zhengyuan Yang, Hanwang Zhang, Zicheng Liu, and Lijuan Wang.
\newblock Disco: Disentangled control for realistic human dance generation.
\newblock In \emph{Proceedings of the IEEE/CVF Conference on Computer Vision and Pattern Recognition}, pp.\  9326--9336, 2024{\natexlab{c}}.

\bibitem[Wang et~al.(2024{\natexlab{d}})Wang, Chen, Ma, Zhou, Huang, Wang, Yang, He, Yu, Yang, et~al.]{wang2024lavie}
Yaohui Wang, Xinyuan Chen, Xin Ma, Shangchen Zhou, Ziqi Huang, Yi~Wang, Ceyuan Yang, Yinan He, Jiashuo Yu, Peiqing Yang, et~al.
\newblock Lavie: High-quality video generation with cascaded latent diffusion models.
\newblock \emph{International Journal of Computer Vision}, pp.\  1--20, 2024{\natexlab{d}}.

\bibitem[Wang et~al.(2004)Wang, Bovik, Sheikh, and Simoncelli]{wang2004image}
Zhou Wang, Alan~C Bovik, Hamid~R Sheikh, and Eero~P Simoncelli.
\newblock Image quality assessment: from error visibility to structural similarity.
\newblock \emph{IEEE Transactions on Image Processing}, 13\penalty0 (4):\penalty0 600--612, 2004.

\bibitem[Wu et~al.(2024)Wu, Jiang, Liu, Yuan, Bai, and Bai]{wu2024general}
Junfeng Wu, Yi~Jiang, Qihao Liu, Zehuan Yuan, Xiang Bai, and Song Bai.
\newblock General object foundation model for images and videos at scale.
\newblock In \emph{Proceedings of the IEEE/CVF Conference on Computer Vision and Pattern Recognition}, pp.\  3783--3795, 2024.

\bibitem[Xing et~al.(2024)Xing, Xia, Zhang, Chen, Yu, Liu, Liu, Wang, Shan, and Wong]{xing2024dynamicrafter}
Jinbo Xing, Menghan Xia, Yong Zhang, Haoxin Chen, Wangbo Yu, Hanyuan Liu, Gongye Liu, Xintao Wang, Ying Shan, and Tien-Tsin Wong.
\newblock Dynamicrafter: Animating open-domain images with video diffusion priors.
\newblock In \emph{European Conference on Computer Vision}, pp.\  399--417. Springer, 2024.

\bibitem[Xiong et~al.(2018)Xiong, Luo, Ma, Liu, and Luo]{xiong2018learning}
Wei Xiong, Wenhan Luo, Lin Ma, Wei Liu, and Jiebo Luo.
\newblock Learning to generate time-lapse videos using multi-stage dynamic generative adversarial networks.
\newblock In \emph{Proceedings of the IEEE Conference on Computer Vision and Pattern Recognition}, pp.\  2364--2373, 2018.

\bibitem[Xu et~al.(2023)Xu, Zhang, Peng, Ma, Jesslen, Ji, Hu, Zhang, Liu, Wang, et~al.]{xu2023animal3d}
Jiacong Xu, Yi~Zhang, Jiawei Peng, Wufei Ma, Artur Jesslen, Pengliang Ji, Qixin Hu, Jiehua Zhang, Qihao Liu, Jiahao Wang, et~al.
\newblock Animal3d: A comprehensive dataset of 3d animal pose and shape.
\newblock In \emph{Proceedings of the IEEE/CVF International Conference on Computer Vision}, pp.\  9099--9109, 2023.

\bibitem[Xu et~al.(2024)Xu, Zhang, Liew, Yan, Liu, Zhang, Feng, and Shou]{xu2024magicanimate}
Zhongcong Xu, Jianfeng Zhang, Jun~Hao Liew, Hanshu Yan, Jia-Wei Liu, Chenxu Zhang, Jiashi Feng, and Mike~Zheng Shou.
\newblock Magicanimate: Temporally consistent human image animation using diffusion model.
\newblock In \emph{Proceedings of the IEEE/CVF Conference on Computer Vision and Pattern Recognition}, pp.\  1481--1490, 2024.

\bibitem[Yang et~al.(2024{\natexlab{a}})Yang, Yang, Hui, Zheng, Yu, Zhou, Li, Li, Liu, Huang, et~al.]{yang2024qwen2}
An~Yang, Baosong Yang, Binyuan Hui, Bo~Zheng, Bowen Yu, Chang Zhou, Chengpeng Li, Chengyuan Li, Dayiheng Liu, Fei Huang, et~al.
\newblock Qwen2 technical report.
\newblock \emph{arXiv preprint arXiv:2407.10671}, 2024{\natexlab{a}}.

\bibitem[Yang et~al.(2024{\natexlab{b}})Yang, Kang, Huang, Zhao, Xu, Feng, and Zhao]{depth_anything_v2}
Lihe Yang, Bingyi Kang, Zilong Huang, Zhen Zhao, Xiaogang Xu, Jiashi Feng, and Hengshuang Zhao.
\newblock Depth anything v2.
\newblock \emph{arXiv:2406.09414}, 2024{\natexlab{b}}.

\bibitem[Yang et~al.(2023)Yang, Du, Dai, Schuurmans, Tenenbaum, and Abbeel]{yang2023probabilistic}
Mengjiao Yang, Yilun Du, Bo~Dai, Dale Schuurmans, Joshua~B Tenenbaum, and Pieter Abbeel.
\newblock Probabilistic adaptation of text-to-video models.
\newblock \emph{arXiv preprint arXiv:2306.01872}, 2023.

\bibitem[Yu et~al.(2023)Yu, He, Deng, Shen, and Chen]{yu2023convolutions}
Qihang Yu, Ju~He, Xueqing Deng, Xiaohui Shen, and Liang-Chieh Chen.
\newblock Convolutions die hard: Open-vocabulary segmentation with single frozen convolutional clip.
\newblock \emph{Advances in Neural Information Processing Systems}, 36:\penalty0 32215--32234, 2023.

\bibitem[Zeng et~al.(2024)Zeng, Wei, Zheng, Zou, Wei, Zhang, and Li]{zeng2024make}
Yan Zeng, Guoqiang Wei, Jiani Zheng, Jiaxin Zou, Yang Wei, Yuchen Zhang, and Hang Li.
\newblock Make pixels dance: High-dynamic video generation.
\newblock In \emph{Proceedings of the IEEE/CVF Conference on Computer Vision and Pattern Recognition}, pp.\  8850--8860, 2024.

\bibitem[Zhang et~al.(2024{\natexlab{a}})Zhang, Wu, Liu, Zhao, Ran, Gu, Gao, and Shou]{zhang2024show}
David~Junhao Zhang, Jay~Zhangjie Wu, Jia-Wei Liu, Rui Zhao, Lingmin Ran, Yuchao Gu, Difei Gao, and Mike~Zheng Shou.
\newblock Show-1: Marrying pixel and latent diffusion models for text-to-video generation.
\newblock \emph{International Journal of Computer Vision}, pp.\  1--15, 2024{\natexlab{a}}.

\bibitem[Zhang et~al.(2020)Zhang, Xu, Liu, Wang, Wu, Liu, and Jiang]{zhang2020dtvnet}
Jiangning Zhang, Chao Xu, Liang Liu, Mengmeng Wang, Xia Wu, Yong Liu, and Yunliang Jiang.
\newblock Dtvnet: Dynamic time-lapse video generation via single still image.
\newblock In \emph{Computer Vision--ECCV 2020: 16th European Conference, Glasgow, UK, August 23--28, 2020, Proceedings, Part V 16}, pp.\  300--315. Springer, 2020.

\bibitem[Zhang et~al.(2023{\natexlab{a}})Zhang, Rao, and Agrawala]{zhang2023adding}
Lvmin Zhang, Anyi Rao, and Maneesh Agrawala.
\newblock Adding conditional control to text-to-image diffusion models.
\newblock In \emph{Proceedings of the IEEE/CVF International Conference on Computer Vision}, pp.\  3836--3847, 2023{\natexlab{a}}.

\bibitem[Zhang et~al.(2023{\natexlab{b}})Zhang, Guo, Pan, Cai, Hong, Li, Yang, and Liu]{zhang2023remodiffuse}
Mingyuan Zhang, Xinying Guo, Liang Pan, Zhongang Cai, Fangzhou Hong, Huirong Li, Lei Yang, and Ziwei Liu.
\newblock Remodiffuse: Retrieval-augmented motion diffusion model.
\newblock In \emph{Proceedings of the IEEE/CVF International Conference on Computer Vision}, pp.\  364--373, 2023{\natexlab{b}}.

\bibitem[Zhang et~al.(2024{\natexlab{b}})Zhang, Cai, Pan, Hong, Guo, Yang, and Liu]{zhang2024motiondiffuse}
Mingyuan Zhang, Zhongang Cai, Liang Pan, Fangzhou Hong, Xinying Guo, Lei Yang, and Ziwei Liu.
\newblock Motiondiffuse: Text-driven human motion generation with diffusion model.
\newblock \emph{IEEE transactions on pattern analysis and machine intelligence}, 46\penalty0 (6):\penalty0 4115--4128, 2024{\natexlab{b}}.

\bibitem[Zhang et~al.(2018)Zhang, Isola, Efros, Shechtman, and Wang]{zhang2018unreasonable}
Richard Zhang, Phillip Isola, Alexei~A Efros, Eli Shechtman, and Oliver Wang.
\newblock The unreasonable effectiveness of deep features as a perceptual metric.
\newblock In \emph{Proceedings of the IEEE Conference on Computer Vision and Pattern Recognition}, pp.\  586--595, 2018.

\bibitem[Zhou et~al.(2022)Zhou, Wang, Yan, Lv, Zhu, and Feng]{zhou2022magicvideo}
Daquan Zhou, Weimin Wang, Hanshu Yan, Weiwei Lv, Yizhe Zhu, and Jiashi Feng.
\newblock Magicvideo: Efficient video generation with latent diffusion models.
\newblock \emph{arXiv preprint arXiv:2211.11018}, 2022.

\bibitem[Zhu et~al.(2024)Zhu, Chen, Dai, Su, Xu, Cao, Yao, Zhu, and Zhu]{zhu2024champ}
Shenhao Zhu, Junming~Leo Chen, Zuozhuo Dai, Qingkun Su, Yinghui Xu, Xun Cao, Yao Yao, Hao Zhu, and Siyu Zhu.
\newblock Champ: Controllable and consistent human image animation with 3d parametric guidance.
\newblock \emph{arXiv preprint arXiv:2403.14781}, 2024.

\bibitem[Zuffi et~al.(2017)Zuffi, Kanazawa, Jacobs, and Black]{zuffi20173d}
Silvia Zuffi, Angjoo Kanazawa, David~W. Jacobs, and Michael~J. Black.
\newblock 3d menagerie: Modeling the 3d shape and pose of animals.
\newblock In \emph{Proceedings IEEE Conference on Computer Vision and Pattern Recognition (CVPR)}, pp.\  5524--5532, 2017.

\bibitem[Zuffi et~al.(2024)Zuffi, Mellbin, Li, Hoeschle, Kjellström, Polikovsky, Hernlund, and Black]{Zuffi:CVPR:2024}
Silvia Zuffi, Ylva Mellbin, Ci~Li, Markus Hoeschle, Hedvig Kjellström, Senya Polikovsky, Elin Hernlund, and Michael~J. Black.
\newblock {VAREN}: Very accurate and realistic equine network.
\newblock In \emph{IEEE/CVF Conference on Computer Vision and Pattern Recognition (CVPR)}, 2024.

\end{thebibliography}
\bibliographystyle{tmlr}

\clearpage
\setcounter{page}{1}
\appendix

\section*{Appendix}
\label{sec:appendix}

\noindent The supplementary material includes the following additional information.

\begin{itemize}
    \item Sec.~\ref{sec:supp:PPPM} provides the architectural details for \motionmodelname.
    \item Sec.~\ref{sec:dataset} provides additional details for our annotated dataset.
    \item Sec.~\ref{sec:supp:abl} provides additional ablation studies omitted from the main paper.
    \item Sec.~\ref{sec:supp:limitation} discusses the limitations of our method.
    \item Sec.~\ref{sec:impacts} discusses the societal impacts of our method.
\end{itemize}

\noindent We also provide the generated videos used in all figures in the main paper, as well as additional videos demonstrating accurate motion control, in the \textit{supplementary videos}.

\section{Architectural details for \motionmodelname}
\label{sec:supp:PPPM}
\begin{wrapfigure}{r}{0.45\textwidth}
    \centering
    \vspace{-6mm}
    \includegraphics[width=0.8\linewidth]{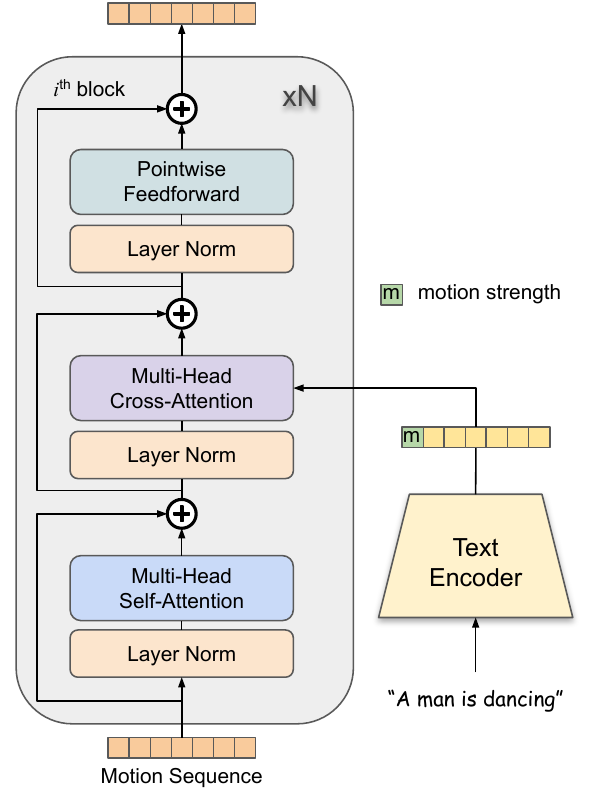}
    \caption{\textbf{Architecture of the Parameterized Motion Prior model}}
    \label{fig:pppm}
    \vspace{-15mm}
\end{wrapfigure}
To optimize the 3D motion sequence extracted from the coarse generated video, we propose the Parameterized Motion Prior model (\motionmodelname).
As shown in Fig.~\ref{fig:pppm}, \motionmodelname utilizes a transformer architecture with self-attention, cross-attention, and feedforward layers as its backbone. 
It takes the parameterized motion sequence of the coarse video as input and optimizes it based on the input text prompt and motion strength.
More specifically, \motionmodelname is trained as a single shared model across all categories. It directly predicts the corrected pose from the perturbed input, and we supervise it with an MSE loss between the predicted pose and the ground-truth pose.

\section{Dataset}
\label{sec:dataset}

Both the motion-conditioned video generation model and the Parameterized Motion Prior model (\motionmodelname) need to be fine-tuned (trained) on a small yet high-quality video dataset with object-centric annotations.
Existing large-scale video datasets~\citep{bain2021frozen,chen2024panda} mainly provide text-image pairs without detailed object-centric annotations.
To address this limitation, we use a suite of off-the-shelf models across various tasks to generate detailed 2D and 3D object-centric annotations. 
We annotate a total of 20K videos from the Panda-70M~\citep{chen2024panda} dataset, with approximately 55\% human videos, 15\% animal videos, and 30\% general-object videos.
For each video, we provide frame-wise 2D bounding boxes, semantic masks, depth estimation maps, and 3D parametric mesh reconstructions for detected humans and animals.
The details are outlined below.

\noindent\textbf{High-Quality Motion Videos Filtering.}
To start with, we use LLMs~\citep{yang2024qwen2} and an open-vocabulary segmentation model~\citep{yu2023convolutions} to curate high-quality motion videos. 
Specifically, LLM filters videos with evident motion based on their captions. 
Then, for each selected video, we equally sample 10 frames and apply the segmentation model to identify humans and animals. 
We evaluate each frame based on the predicted mask size and mask count.
Then we retain videos where humans or animals occupy a significant portion of the frame and where the count of humans does not exceed five in each frame. 

\noindent\textbf{Object Detection and Depth Estimation.}
Based on the captions of videos, we identify the objects mentioned and detect their bounding boxes~\citep{varghese2024yolov8} and instance masks~\citep{yu2023convolutions}. 
We also apply Depth Anything V2~\citep{depth_anything_v2} to generate the depth maps of each frame.

\noindent\textbf{Human Videos Annotation.}
For videos containing humans, we focus on extracting 2D instance segmentation masks, 2D part masks, 2D face keypoints, 3D body pose and shape, and 3D hand pose. 
We begin by using YOLO-V8~\citep{varghese2024yolov8} to segment all humans in each frame, providing accurate human masks. 
Next, we apply a state-of-the-art face keypoint detector, RTMPose~\citep{jiang2023rtmpose}, to predict facial keypoints for each detected human.
Simultaneously, we use 4D-Human~\citep{goel2023humans} and HaMeR~\citep{pavlakos2024reconstructing} to estimate the 3D body and hand meshes.
The resulting SMPL (body mesh)~\citep{loper2023smpl} and MANO (hand mesh)~\citep{romero2022embodied} parameters are then fit into a unified SMPL-X~\citep{pavlakos2019expressive} representation, which contains both human body and hand meshes.
We then project the 3D SMPL-X human mesh onto 2D to obtain part masks, as each vertex in the SMPL-X mesh is labeled by body part.
Finally, we project the face keypoints and 3D human mesh onto the instance mask, allowing us to compute the overlap between the projected keypoints, projected human mask, and detected 2D human mask. 
This overlap is quantified using an IoU score, which is used to filter out annotations with high errors.
As a result, for each video, we obtain annotations including human instance masks, 2D facial keypoints, 3D SMPL-X meshes for the body and hands, and 2D part-level segmentation masks.

\noindent\textbf{Animal Videos Annotation.}
We start by using Grounded SAM 2~\citep{kirillov2023segany, ravi2024sam2segmentimages, ren2024grounded} to segment animal masks in each frame. 
Next, we apply a state-of-the-art camera estimation algorithm, VGGSfM~\citep{wang2024vggsfm}, to optimize the camera's intrinsic and extrinsic parameters across the video.
To ensure a reliable camera estimate, we set thresholds on mean projection errors and mean track lengths, filtering out videos that do not meet these criteria.
We then use AnimalAvatar~\citep{Sabathier2024AnimalAR} initialized with Animal3D~\citep{xu2023animal3d} to fit SMAL parameters.
Each video is divided into segments of 10 consecutive frames, and AnimalAvatar is applied to each segment independently.
This strategy helps mitigate the impact of outliers in camera predictions on the overall optimization quality.
To ensure the accuracy of SMAL fitting, we impose thresholds on IoU and PSNR~\citep{hore2010image}, filtering out video segments that do not meet our accuracy standards.
Once accurate SMAL fittings are obtained, we follow a similar pipeline to extract the desired annotations as used in human cases.

\section{Ablation Study}
\label{sec:supp:abl}
Herein, we conduct additional ablation studies to verify the effectiveness of the proposed designs.

\textbf{Quantitative Evaluation Regarding Inference Efficiency.}
\begin{table}
\vspace{-3mm}
  \centering
  \caption{{\small \textbf{Quantitative evaluation regarding inference efficiency.} We reduce Stage 1 compute time from 36 seconds to 8 seconds, while maintaining comparable final video quality.}}
  \label{tab:efficient}
  \resizebox{0.85\linewidth}{!}{%
    \begin{tabular}{l|cccc}
       & {\bf Overall Preference} & {\bf Visual Quality} & {\bf Motion Consistency} & {\bf Amount of Motion} \\
      \midrule
      \modelname-SVD (full model)	& 52.60\% & 50.27\% & 51.87\% & 54.27\% \\
      \modelname-SVD (efficient model) & 47.40\% & 49.73\% & 48.13\% & 45.73\% \\
    \end{tabular}%
  }
  \vspace{-3mm}
\end{table}

Stage 1 can be performed with reduced computational overhead while maintaining similar results. To quantify this, we conducted experiments on $500$ video pairs using the same input images but with varying settings: lower resolution (1:4), fewer frames (8 vs. 16), and fewer denoising steps (32 vs. 50). The efficient model reduced Stage 1 compute time from 36 seconds to 8 seconds. We ran a human preference study on Amazon MTurk, comparing the full and efficient versions for each video, with results provided in Tab.~\ref{tab:efficient}. The results demonstrate that the efficient model achieves similar performance to the full model in visual quality and motion consistency, with only a slight drop in the amount of motion, which is likely due to the reduced number of frames. This confirms that Stage 1 can be significantly accelerated with minimal impact on perceived quality.

\textbf{Parameterized Motion Prior Model.} We briefly discuss the quantitative improvements of the proposed \motionmodelname in the main paper and provide additional experimental details and results here.
To further demonstrate the improvements of \motionmodelname, we conducted an additional user study on Amazon MTurk comparing videos generated with and without \motionmodelname. 
Unlike our previous study, which compared our method with SVD, this evaluation focuses on object consistency and motion consistency. 
We also report the percentage of videos containing incorrect human or animal structures (\ie, the morphological failure rate).
We evaluate 500 video pairs, each rated by three different users, resulting in 1,500 total evaluations. The results are presented in Tab.~\ref{supp:tab:user} of the main paper.
By optimizing with a parametric 3D mesh, our approach significantly reduces incorrect human and animal structures, leading to substantial improvements in object and motion consistency.

\begin{figure}
    \centering
    \includegraphics[width=0.75\linewidth]{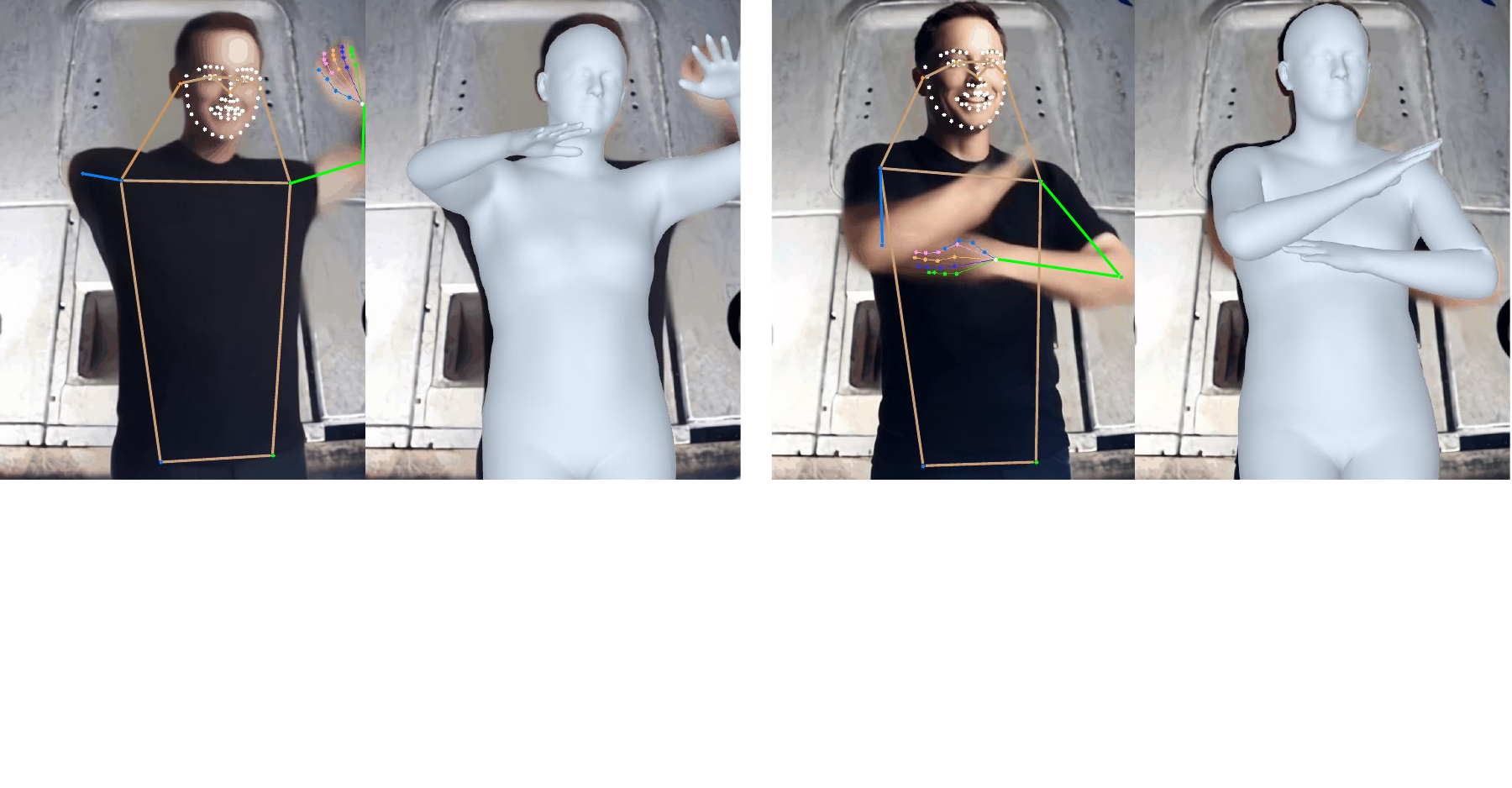}
    \vspace{-22mm}
    \caption{
    \textbf{The parametric 3D mesh serves as an effective object-level prior,} ensuring complete human body structures in the coarse video generated during the first stage. In the left two images, the human keypoint model fails to detect the missing right hand, which is accurately ``recovered" by the parametric human mesh model. In the right two images, the human mesh model provides a more accurate prior for both blurred hands.
    }
    \vspace{-2mm}
    \label{fig:abl_smpl_kp}
\end{figure}

\noindent{\textbf{Parametric 3D Mesh.}}
Previous human image animation models mainly rely on 2D pose sequences for each frame to provide motion information.
However, this approach is not optimal for general video generation. 
As shown in Fig.~\ref{fig:abl_smpl_kp}, we compare the results of using a parametric human mesh model~\citep{loper2023smpl} versus a human keypoint model~\citep{sun2019deep}.
Our findings indicate that the human mesh model provides a robust object-level prior, which significantly benefits general video generation.
Specifically, current video generation models often misinterpret the structure of humans and animals, occasionally producing unrealistic results, such as a man with three arms, an example of morphological failure.
This problem becomes more serious in complex motion generation. 
However, incorporating human and animal priors from 3D mesh models substantially mitigates these structural inaccuracies, enabling more accurate representations of targets.

\begin{figure}[ht]
    \centering
    \includegraphics[width=0.98\linewidth]{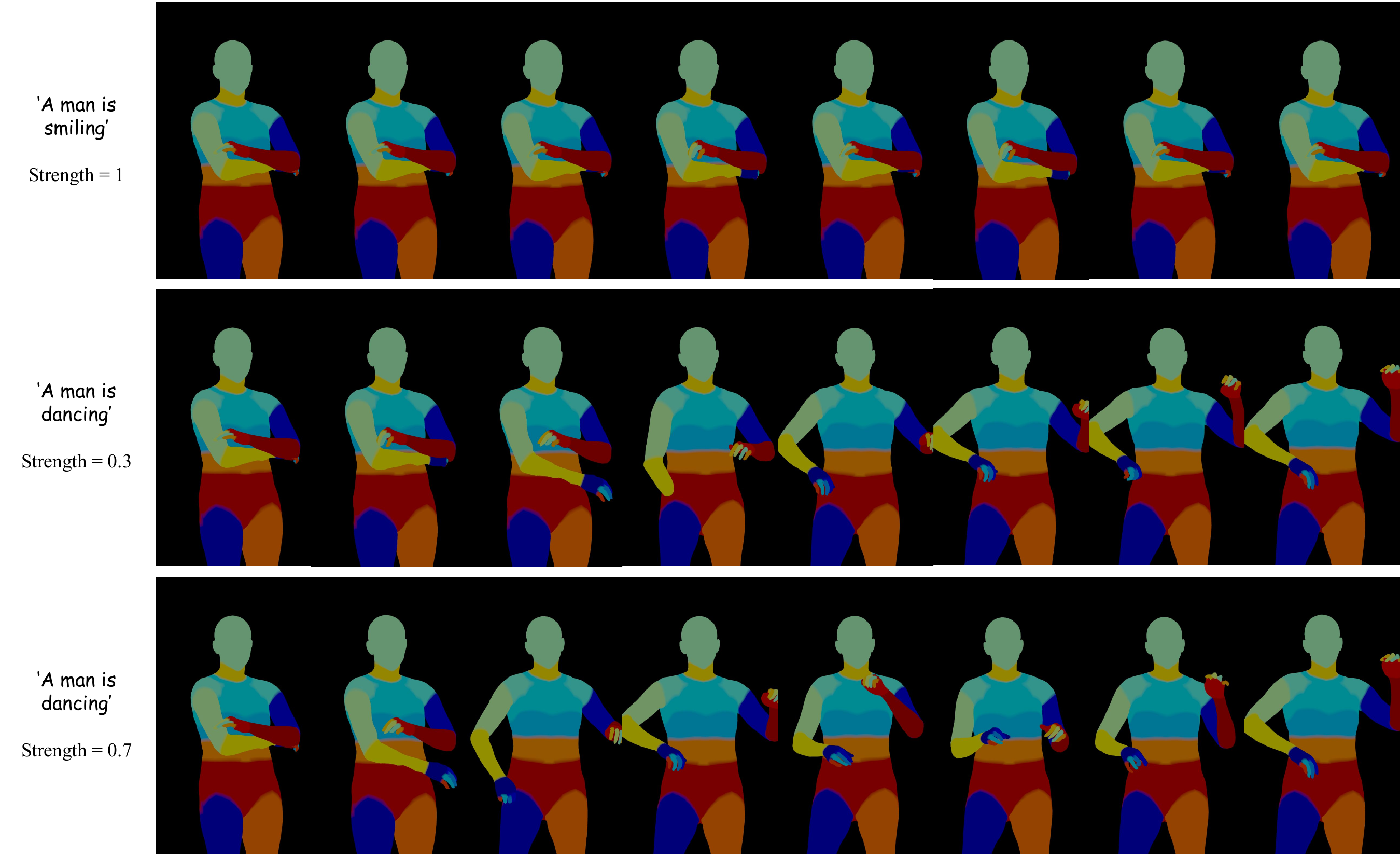}
    \caption{\textbf{Text prompt and motion strength.} We show projected motion sequences generated using different text prompts and motion strength parameters. The text prompt helps define the motion style, while the motion strength controls the speed and complexity of the generated motion. For example, a motion strength of 0.3 results in a simple, direct trajectory, whereas a strength of 0.7 produces a more dynamic and complex motion path. The same target pose is used for the second and third rows to highlight the effect of motion strength.}
    \label{fig:MotStr}
\end{figure}

\noindent{\textbf{Text Prompt and Motion Strength.}} Generating videos from a single image introduces significant ambiguity. To reduce this, we incorporate additional conditioning using a text prompt and a motion strength parameter. Specifically, the text prompt defines the intended motion type, while motion strength controls the speed and complexity of motion within the video.
In our experiments, we observe that varying motion strength with the same target pose leads to different motion trajectories. For instance, when moving a hand from point A to B, a video generated with low motion strength results in a direct, simple movement. In contrast, higher motion strength produces a more dynamic and complex trajectory, though it still reaches the same final pose at B.
An illustrative example is provided in Fig.~\ref{fig:MotStr}.

\begin{table}[t]
\centering
\caption{\textbf{Ablation on confidence score.}
We evaluate three parameter configurations corresponding to the $\{$full motion sequence, polygon target pose, empty$\}$ conditions: (1, 0.5, 0), (0.8, 0.5, 0.2), and (3, 2, 1). The results demonstrate that performance is highly robust to the specific choice of confidence values.
}
\label{tab:conf_score_ablation}
\resizebox{0.5\linewidth}{!}{%
\begin{tabular}{l|lccc}
\textbf{Model} & \textbf{Setting} & \textbf{SSIM} & \textbf{PSNR} & \textbf{LPIPS} \\
\midrule
\textit{VividPose}  & -       & \textit{0.758} & \textit{29.83} & \textit{0.261} \\
Ours & (1, 0.5, 0)               & 0.864          & 30.08          & 0.210          \\
Ours & (0.8, 0.5, 0.2)           & 0.851          & 30.10          & 0.214          \\
Ours & (3, 2, 1)                 & 0.873          & 30.07          & 0.217          \\
\end{tabular}}
\end{table}
\noindent{\textbf{Ablation on the Confidence Score for Conditioning.}}
We use the confidence map as simple markers to indicate whether a given spatial location contains motion supervision. This is consistent with common practice in prior work (\eg, Emu Video~\citep{girdhar2023emu}), where a binary mask is used purely to denote the presence or absence of conditioning signals. In our formulation, assigning different numeric values (\eg, 1 for a full motion sequence, 0.5 for a target pose, and 0 for empty) does not introduce additional learnable meaning. Rather, these values serve as soft indicators that allow us to unify different conditioning types within a single representation, and the model does not depend on their exact magnitudes. We verify this with an ablation in Tab.~\ref{tab:conf_score_ablation}, where we evaluate three parameter sets, \ie, (1, 0.5, 0), (0.8, 0.5, 0.2), and (3, 2, 1), for \{full motion sequence, polygon target pose, empty\}. The results show that performance is highly robust to the specific choice of confidence values. Given this insensitivity, learning an additional confidence-prediction head is unlikely to provide meaningful gains, while it would introduce extra complexity; therefore, we adopt this simple, standard, and effective marking strategy.

\section{Limitations}
\label{sec:supp:limitation}

The proposed method has several remaining limitations.
\emph{First}, it relies on parametric 3D mesh models, requiring multiple off-the-shelf models for different object categories, though it adds only 5 seconds to the total inference time. 
Recent advances in 3D modeling, such as encoding 3D priors of general objects within a single diffusion model~\citep{liu2024direct}, are paving the way for more general, efficient models that can be seamlessly integrated into our pipeline for high-quality video generation.
\emph{Second}, the model still struggles to generate high-quality details such as fingers and hands.
\emph{Finally}, while \motionmodelname can generate realistic motion sequences beyond 32 frames, our current implementation, based on vanilla SVD, is limited by memory constraints (80GB RAM). However, recent methods have demonstrated longer video generation ability using pretrained diffusion models~\citep{chen2023seine}. Exploring long-video generation with 3D knowledge remains future work.

\section{Statement of Broader Impact}
\label{sec:impacts}
The proposed \modelname has the potential to facilitate numerous fields through its advanced video generation capabilities. In the realm of creative industries, \modelname can enhance the efficiency and creativity of artists and designers by generating high-fidelity videos. The high-quality generated videos can also contribute to research on synthetic datasets by creating realistic videos, aiding in reducing the annotations required for training vision models. However, with these advancements come ethical considerations, such as the risk of generating deepfakes or other malicious content. It is thus crucial to implement safeguards to minimize potential harms.

\end{document}